\documentclass[letterpaper, 10 pt, conference]{ieeeconf}  % Comment this line out if you need a4paper

\IEEEoverridecommandlockouts                              % This command is only needed if 
\UseRawInputEncoding  % supress invalid UTF-8 byte sequence

\usepackage{graphics} % for pdf, bitmapped graphics files
\usepackage{epsfig} % for postscript graphics files
\usepackage{mathptmx} % assumes new font selection scheme installed
\usepackage{times} % assumes new font selection scheme installed
\usepackage{amsmath} % assumes amsmath package installed
\usepackage{amssymb}  % assumes amsmath package installed, checkmark symbol
\usepackage{booktabs} % typeset tables
\usepackage{bm} %boldsymbol
\usepackage{subfig}
\usepackage{caption}
\usepackage{relsize}
\usepackage{graphicx}
\usepackage{wrapfig}
\usepackage{lscape}
\usepackage{rotating}
\usepackage{epstopdf}
\usepackage{rotating}
\usepackage{scalerel}

\usepackage{multirow}
\usepackage{url}

\usepackage{cite}
\title{\LARGE \bf
Sitting, Standing and Walking Control of the Series-Parallel Hybrid Recupera-Reha Exoskeleton}

\author{Ibrahim Tijjani$^{1}$, Rohit Kumar$^{1,*}$, Melya Boukheddimi$^{1,*}$, Mathias Trampler$^{1}$, Shivesh Kumar$^{1,2}$ \\ and Frank Kirchner$^{1,3}$% <-this % stops a space
\thanks{$^{1}$ Robotics Innovation Center, German Research Center for Artificial Intelligence (DFKI GmbH), 28359 Bremen, Germany.}%
\thanks{$^{2}$ Division of Dynamics, Department of Mechanics \& Maritime Sciences, Chalmers University of Technology, Gothenburg, Sweden}%
\thanks{$^{3}$ AG Robotik, University of Bremen, Bremen, Germany.}%
\thanks{$^*$ Both authors equally contributed to this work.}
\thanks{This work was supported by the M-RoCK (FKZ 01IW21002), VeryHuman (FKZ 01IW20004), and CoEx (01IW24008) projects funded with federal funds from the Federal Ministry of Economic Affairs and Energy (BMWi).}
\thanks{Corresponding author: ibrahim.tijjani@dfki.de}
}

\begin{document}

\newcommand{\mvec}[1]{\bm{#1}}
\newcommand{\vc}[1]{\mathbf{\mathbf{#1}}}

%%%%%%%%%%%%%%%%%%%%%%%%%%%%%%%%%%%%%%%%%

% \newcommand{\q}{q}
% \newcommand{\dq}{\dot{\q}}
% \newcommand{\ddq}{\ddot{\q}}
\newcommand{\q}{\mathbf{q}}
\newcommand{\dq}{\mathbf{\dot{\q}}}
\newcommand{\ddq}{\mathbf{\ddot{\q}}}

%%%%%%%%%%%%%%%%%%%%%%%%%%%%%%%%%%%%%%%

\newcommand{\Mass}{\mathbf{M}}
\newcommand{\Bias}{\mathbf{b}}
\newcommand{\Gravity}{\mathbf{g}}
\newcommand{\Force}{\mathbf{\lambda}}
\newcommand{\Torque}{\mathbf{\tau}}
\newcommand{\Jac}{\mathbf{J}}

%%%%%%%%%%%%%%%%%%%%%%%%%%%%%%%%%%%%%%%%%
\newcommand{\BIN}{\begin{bmatrix}}
\newcommand{\BOUT}{\end{bmatrix}}

%%%%%%%%%%%%%%%%%%%%%%%%%%%%%%%%%%%%%%%%%
\newcommand{\sref}[1]{Sec~\ref{#1}}
\newcommand{\eref}[1]{(\ref{#1})}
\newcommand{\fref}[1]{Fig.~\ref{#1}}
\newcommand{\tref}[1]{Table~\ref{#1}}

%%%%%%%%%%%%%%%%%%%%%%%%%%%%%%%%%%%%%%%%
\newcommand{\state}{\mathbf{x}}
\newcommand{\ctrl}{\mathbf{u}}
\newcommand{\dynsys}{\mathbf{f}}

%%%%%%%%%%%%%%%%%%%%%%%%%%%%%%%%%%%%%%%
\newcommand{\qTr}{\underline{\q}}
\newcommand{\dqTr}{\underline{\dq}}
\newcommand{\ddqTr}{\underline{\ddq}}
\newcommand{\TorqueTr}{\underline{\Torque}}

%%%%%%%%%%%%%%%%%%%%%%%%%%%%%%%%%%%%%%
\newcommand{\costl}{l}
\newcommand{\dts}{\Delta t_s}
\newcommand{\st}{\text{subject to}}

\maketitle
\thispagestyle{empty}
\pagestyle{empty}

\begin{abstract}
%This paper presents advances in the functionalities of the full-body Recupera-Reha exoskeleton robot, primarily designed to support upper body weight in assistive upper limb rehabilitation. The robot design consists of various linear actuators working in a parallel configuration to better transfer the load from the upper body to the ground. This has resulted in a series-parallel hybrid design with a large number of kinematic loops which is challenging to model and control.
%To address the challenge of controlling the exoskeleton lower extremity for walking purposes, we demonstrate the power of optimal control to generate feasible motions such as sitting, standing and static walking, then adapt these motions on the robot.
%Our modeling and control approach takes advantage of a serial abstraction of the model to solve the optimal control problem efficiently and then uses a full series-parallel hybrid model which respects all the loop closure constraints for generating final actuator commands.
%The experimental results demonstrate the versatility of our control approach in generating desired motions for the exoskeleton.

This paper presents advancements in the functionalities of the Recupera-Reha lower extremity exoskeleton robot. The exoskeleton features a series-parallel hybrid design characterized by multiple kinematic loops resulting in 148 degrees of freedom in its spanning tree and 102 independent loop closure constraints, which poses significant challenges for modeling and control. To address these challenges, we applied an optimal control approach to generate feasible trajectories such as sitting, standing, and static walking, and tested these trajectories on the exoskeleton robot. 
Our method efficiently solves the optimal control problem using a serial abstraction of the model to generate trajectories. It then utilizes the full series-parallel hybrid model, which takes all the kinematic loop constraints into account to generate the final actuator commands. The experimental results demonstrate the effectiveness of our approach in generating the desired motions for the exoskeleton.
\end{abstract}

\section{Introduction}
\label{sec_introduction}
%Exoskeleton robots have the potential to revolutionize fields such as medical, industrial, military, search and rescue operations by enhancing human capabilities for physically demanding tasks. As the demand for assistive technologies grows, these robots will play a crucial role in boosting productivity and improving quality of life. %Current research aims to optimize exoskeleton functionality and adaptability, focusing on advancements in user control systems, lightweight materials, and intuitive interfaces. The performance of exoskeletons is also heavily influenced by the integration of sensors, feedback mechanisms, and ergonomic design, ensuring smooth interaction between users and the robotic system.
Exoskeleton robot research focuses on developing wearable devices that enhance human capabilities, particularly in mobility and strength. These devices, often inspired by the structure and function of the human body, are designed to augment, assist, support, or rehabilitate individuals with various needs, including those with mobility impairments, injuries, or conditions affecting muscle strength. In recent years, exoskeleton research has gained significant traction due to its potential to address a wide range of needs across various domains of which researchers explore areas such as control algorithms, biomechanics, and human-robot interaction to optimize exoskeleton performance, comfort, and usability. Key goals include improving mobility, enhancing rehabilitation outcomes, and promoting independence and quality of life for users. Overall, exoskeleton research represents a dynamic and multidisciplinary field with significant potential to revolutionize healthcare, industry, military, and everyday life activities by enhancing human capabilities, promoting rehabilitation, and improving overall well-being~\cite{gorgey2018robotic},~\cite{siobhan2021exo}. Ongoing research aims to solve problems in kinematics, dynamics and control formulations, which are essential for generating robot motion and the forces that drive it, to improve performance and fully utilize the exoskeleton technology's potential in various contexts.

\begin{figure}[tpb] %[!htb]
\vspace{0.3cm}
%\hfill
\centering
\subfloat{\includegraphics[width= 0.45\textwidth]{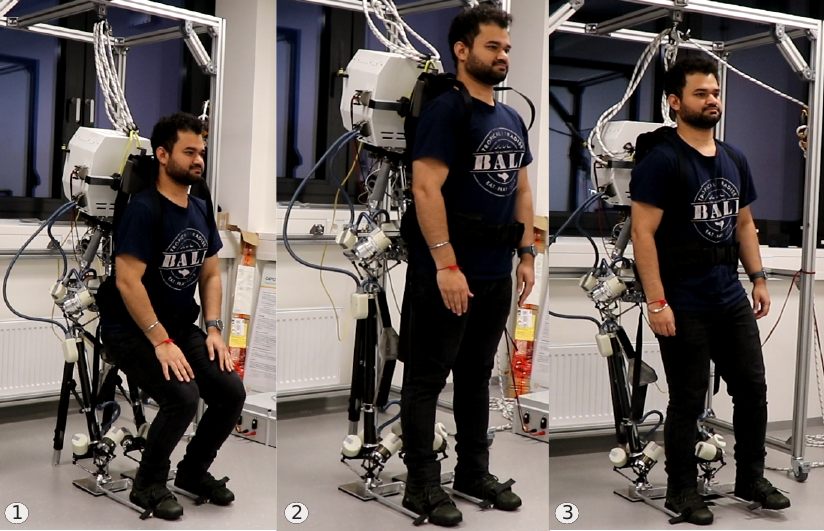}}
\caption{The Recupera-Reha exoskeleton robot in (1) sitting, (2) standing and (3) static walking mode}
\label{fig:1}
\vspace{-0.3cm}
\end{figure}
%\paragraph*{Related works}
\subsection{Related Works}
Over the years, researchers have actively designed wearable exoskeleton robots to enhance ambulatory abilities. While there has been significant progress in the development and implementation of upper-body exoskeletons, commercially deployed lower-body designs remain comparatively fewer (see \cite{tijjani2022survey} for a survey on the design and control of lower extremity exoskeletons). Exoskeleton robots with a series-parallel hybrid structure~\cite{kumar2020survey} face challenges in kinematics and dynamic modeling due to their closed-loop formation and kinematic constraints.
%While we have used classical methods to solve motion equations, we lack a general approach to computing complex dynamic equations from closed-loop multi-body robots.
Modeling tools and software frameworks~\cite{Felis2016rbdl,delp2007opensim,carpentier2019pinocchio} uses the numerical method to solve loop closure constraints in lower-body exoskeletons reported in \cite{meng2023concept,maryam2019human,harib2018feedback} but their solvers may suffer from insufficient accuracy and slow convergence rates.
%However, only a few have decided to use the analytical method, focusing on robot modules already in the solver's database \cite{shivesh2019an,kumar2020survey}.
Recent progress in robot and mechanism modeling has focused on the modular resolution of closed-loop constraints \cite{2017_kumar_mimic,Kum18,kumar2022modular,andreas2022dynamics} which allows using a combination of numerical and analytical methods offering a balanced approach for using already available analytical solutions and numerical resolution of constraints where the analytical solutions are not available.

Fig.~\ref{fig:1} illustrates the Recupera-Reha exoskeleton with a human wearer in three modes: (1) sitting, (2) standing, and (3) static walking. The exoskeleton incorporates a parallel configuration of linear actuators at the spine and knee joints, alongside reported variant of the Almost Spherical Parallel Mechanism (ASPM)~\cite{tijjani2022finding} representing the hip and ankle submechanisms. This complex design yields a series-parallel hybrid structure with numerous kinematic loops, posing challenges in both modeling and control (see~\cite{kumar2020survey} for a survey on such systems). In addition to these complexities, the exoskeleton is equipped with a distinctive user-scalable chair feature that provides relief for users during prolonged standing, thereby enhancing comfort and usability, and ultimately supporting a more adaptable exoskeleton experience.

When it comes to controlling the gait motion of an exoskeleton, whether through position, force, or torque control, it's crucial to carefully define the control algorithm to ensure it adapts effectively to the exoskeleton system and the user's movements. Many exoskeleton designs~\cite{Chaichaowarat2023transformable,tian2024self,yu2023design} rely on classical Proportional, Integral and Derivative (PID) controllers, trajectory tracking, gait pattern generation, motion capture, and Electromyography (EMG) signals from healthy human muscles. These methods are typically used to evaluate the exoskeleton’s assistive performance by ensuring the system responds appropriately to the user's intended motions. Unlike these methods, the outcome of an exoskeleton robot walking in~\cite{Shahrokhshahi2022sample} relies on Model Predictive Control (MPC) based on a quadratic program. This approach specifies kinematic tasks to determine desired joint positions and velocities. In~\cite{harib2018feedback}, Optimal Control (OC) based on direct collocation is used on the full exoskeleton model to produce walking motions. This method formulates OC in a discrete-time domain rather than continuous, simplifying the numerical solution using a non-linear program. However, this method has the tendency to get stuck in local minima and has poorer convergence property in comparision to OC methods based on differential dynamic programming (DDP).

\vspace{-0.1cm}
\subsection{Contributions}
\vspace{-0.05cm}
This research work contributes to advancing the capabilities of the Recupera-Reha exoskeleton. %specifically tailored to support upper-body weight during upper-limb assistive rehabilitation.
First, we extend the kinematic and dynamic analysis to encompass the complete model of the exoskeleton robot, addressing its inherent closed-loop mechanisms through a modular hybrid numerical-analytical approach facilitated by the HyRoDyn~\cite{Kum18, kumar2022modular} software framework. This paper is the first to apply this software approach to a highly complex system (with 148 DOFs in the spanning tree and 102 loop closure constraints). To the best knowledge of the authors, this is one of the most kinematically complex robots shown to be solved with a model-based approach in the literature.
Moreover, by employing DDP based OC method as part of the state-of-the-art (SOTA), we adapted it to handle the specific kinematic and dynamic properties of the exoskeleton, enhancing its functionality and facilitating motions such as sitting, standing, and static walking. Unlike many SOTA methods that remain theoretical or simulated, our work demonstrates the successful application of these methods both in simulation as well as on the real system, validating the practical efficiency of the approach.

\subsection{Organization}
This paper is organized as follows. In Section~\ref{sec_modeling_and_control}, the model formulation and description of the Recupera-Reha exoskeleton is illustrated. In Section~\ref{sec_motion_generation}, the OC problems used for sitting, standing, and static walking are formulated. Section~\ref{sec_experiment_and_results} presents the simulation and experimental results. Finally, Section~\ref{sec_conclusion} offers a discussion and conclusion.

\newcommand{\vect}[1]{\mathbf{#1}}
\section{Modeling of Recupera-Reha System}
\label{sec_modeling_and_control}
The mechanical structure of the Recupera-Reha exoskeleton contains various parallel submechanisms that pose a major challenge in terms of modeling and control. This section describes the modeling approach for the Recupera-Reha system that takes into account the loop closure constraints and enables computationally efficient solutions for the kinematics and dynamics of the robot.
% This section presents the complexity in modeling of Recupera-Reha series-parallel hybrid system. The complexity lies in the fact that the system is composed of multiple loop closure mechanisms which makes it crucial to resolve them in real-time in a computationally efficient manner.

\begin{figure*}[htbp]
	\centering
	\includegraphics[width= 0.99\textwidth]{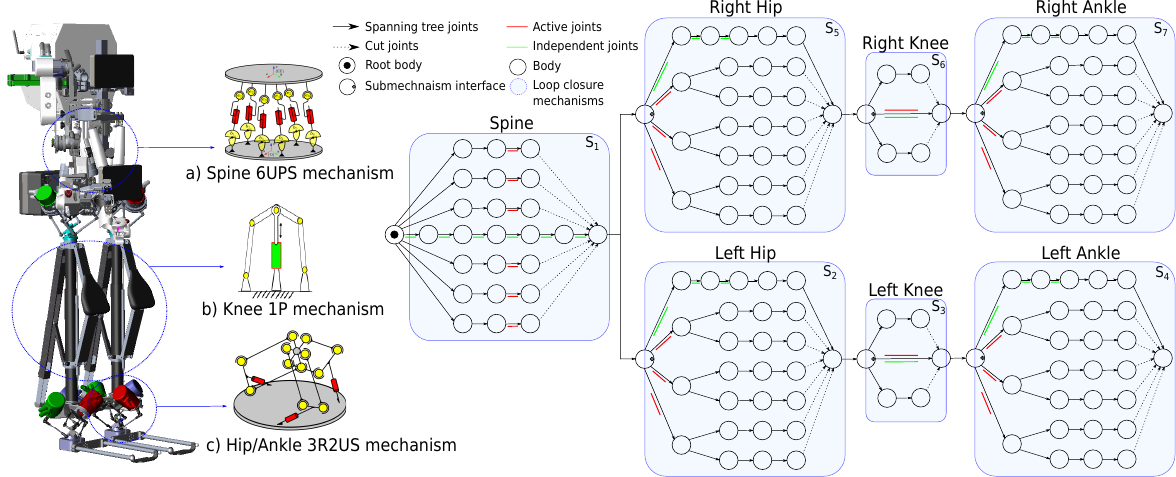}
	\caption{Recupera-Reha system and its topological graph}
	\label{fig:topological}
	\vspace{-0.5cm}
\end{figure*}

\subsection{Modeling of Closed Loops as Explicit Constraints}
Loop closure constraints in rigid body systems are the non-linear constraints on the motion variables. They can be expressed in implicit and explicit form \cite{featherstone2014rigid}. Any rigid body system can be seen as a tree-type system with combination of links and joints. Let $\vect{q}$ be the vector of spanning tree joints and $\vect{y}$ a vector of independent joint variables that define $\vect q$ uniquely (refer to \cite{Kum18} for a detailed description of spanning tree joints and independent joints).
\begin{table}[h]
	\caption{Loop closure constraints \cite{featherstone2014rigid}}
	\label{tab:loop-constriants}
	\begin{center}
		\begin{tabular}{rrrr}
			\toprule
			Type & Position & Velocity & Acceleration\\
			\midrule
			Implicit: & $ \vect{\phi(q)} = \vect{0} $ & $ \vect{K\dot{q}} = \vect{0} $ & $ \vect{K\Ddot{q}} = \vect{k} $ \\
			Explicit: & $ \vect{q} = \vect{\gamma} (\vect{y}) $ & $ \vect{\dot{q}} = \vect{G} \vect{\dot{y}} $ & $ \vect{\Ddot{q}} = \vect{G}\vect{\Ddot{y}} + \vect{g} $\\
			\bottomrule
		\end{tabular}
	\end{center}
	\vspace{-0.5cm}
\end{table}

In Table \ref{tab:loop-constriants}, $\vect{K} = \frac{\partial\vect{\phi}}{\partial\vect{q}}$, $\vect{k} = - \vect{\dot{K}\dot{q}}$, $\vect{G} = \frac{\partial \vect{\gamma}}{\partial \vect{y}}$, and $\vect{g} = \vect{\dot{G}\dot{y}}$. If both implicit and explicit constraints define the same constraint in the system, then it can be deduced that $\vect{\phi} \text{o} \vect{\gamma} = \vect{0}$, $\vect{KG} = \vect{0}$, and $\vect{Kg=k}$. The main focus is on the explicit constraints as it eliminates the possibility of constraint violation. In \cite{kumar2022modular}, a modular approach is described where explicit constraints are derived numerically from implicit constraints in a computationally efficient way.

\subsection{System Description}

This section describes the loop closure mechanisms present in the 34.68 kg Recupera-Reha exoskeleton robot. It has 20 active degree of freedom (DOF) and consists of an electronic backpack, spine, and the lower-limb joints. In Fig. \ref{fig:topological}, the system is represented as a serial composition of 7 submechanisms, all of them are closed-loop mechanisms. It consists of 148 spanning tree joints ($n=148$) of which 20 are independent joints ($m=20$) and 20 are actuated joints ($p=20$). Here, the joints in the mechanism that move independently of other joints are referred to as the independent joints. All the independent joints are shown as green edges and the actuated joints are shown as red edges. The remaining spanning tree joints are passive in nature. In each submechanism, a cut joint approach is used to model loop closures. These cut joints are denoted by dotted lines. The three major loop closure mechanisms are located in the spine, hip and ankle, and in the prismatic knee of the system.

\subsubsection{Spine Mechanism}
The first loop closure submechanism (denoted as $S_1$ in Fig. \ref{fig:topological}) is a stewart platform~\cite{stewart1965a} of type 6U\underline{P}S, actuated by 6 linear actuators.
%The independent joints to control the submechanism are denoted as green edges in $S_1$. 
Here, the spherical joint is considered as the cut joint and each cut joint imposes 3D translation constraint. This submechanism imposes a total of 18 constraints.

\subsubsection{Knee Mechanism}
\begin{figure}[!htbp]
	\centering
	%\framebox{\parbox{3in}}
	%\hspace*{2cm}
	\includegraphics[width= 0.25\textwidth]{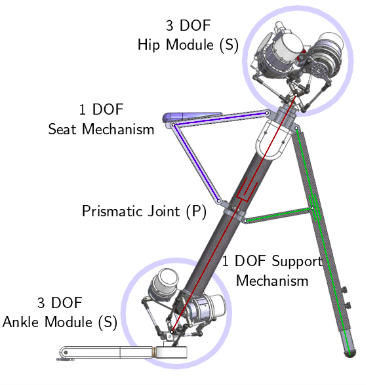}
	\caption{Leg in the Recupera-Reha exoskeleton}
	\label{fig:knee}
	\vspace{-0.3cm}
\end{figure}
%\vspace{-0.3cm}

The second loop closure submechanism is an active-passive knee mechanism actuated by a single prismatic actuator. The knee is designed in such a way that it can be used as a chair to support a person through its passive 1 DOF support and seat mechanism in Fig. \ref{fig:knee}. The support and seat mechanisms considers revolute joints as a cut joint and imposes a total of 6 constraints (3 per planar translation constraints). The submechanism is denoted as $S_3$ and $S_6$ in Fig. \ref{fig:topological}.

\subsubsection{Hip and Ankle Mechanism}
The third submechanism in the system is a multi loop closure mechanism of type 3\underline{R}2US with 3 DOF, present in the hip and ankle of the system. The independent joints are roll, pitch and yaw movements that are used to provide control command to the three actuators present in the submechanisms. The loop closures are defined using spherical cut joints that impose 3D translation constraint. Six cut joints in the submechanism contributes to 18 constraints. The submechanisms are denoted as $S_2$, $S_4$, $S_5$, and $S_7$ in Fig. \ref{fig:topological}.

\subsubsection{Numerical-Analytical Resolution of Loop Closures as Explicit Constraints}
%The Recupera-Reha exoskeleton system is a multi-body system composed of several parallel submechanisms interconnected in a closed loop formation. This closed loop introduces significant complexities in system dynamics and control. To manage these complexities, a numerical-analytical hybrid approach is employed to simplify the model by introducing explicit kinematic constraints, referred to as loop closure constraints, which break the loops and facilitate efficient control. This approach builds upon our previous work \cite{kumar2022modular}, where a similar methodology was applied to more general systems. In this scenario, an analytical approach was exclusively used to resolve the spine mechanism, benefiting from the availability of symbolic expressions which made the process straightforward. However, tackling the hip and ankle mechanisms required a significant investment of expert knowledge and time to derive the necessary analytical expressions, as they lacked existing symbolic representations. The adoption of a numerical-analytical approach efficiently resolved the model complexity by introducing cut joints as constraints to break the kinematic loops. This approach proved highly advantageous, enabling the explicit resolution of loop closures, significantly simplifying the model and ultimately saving several weeks of computational time compared to alternative methods. The explicit constraints can be written as
The numerical-analytical hybrid approach is used to model the loop closures as explicit constraint (in Table \ref{tab:loop-constriants}) for the system, building upon our previous work \cite{kumar2022modular}. It leverages the modularity of the robot's design by applying analytical loop closure for known submechanisms, while using numerical loop closure for submechanisms lacking analytical solutions. In this scenario, an analytical approach was exclusively applied to resolve loop closures in the spine mechanism, benefiting from the availability of symbolic expressions. However, tackling the hip and ankle mechanisms required a significant investment of expert knowledge and time to derive the necessary analytical expressions, as they lacked existing symbolic representations. Adopting a numerical-analytical approach proved highly advantageous, enabling the explicit resolution of loop closures and ultimately saving several weeks of computational time compared to alternative methods. The explicit constraints can be written as
\begin{equation}
\vect{\gamma}=
\left[
\begin{array}{ccccccc}
{\gamma^T_{1,a}} & {\gamma^T_{2,n}} &  {\gamma^T_{3,n}} &  {\gamma^T_{4,n}}& {\gamma^T_{5,n}} &  {\gamma^T_{6,n}} &  {\gamma^T_{7,n}}
\end{array}
\right] ^{T}
\label{eqn_recupera_leg_gamma}
\end{equation}
\begin{equation}
\vect{G}=\text{diag}
\left(\vect{G}_{1,a},\vect{G}_{2,n},\vect {G}_{3,n},\vect {G}_{4,n},\vect {G}_{5,n},\vect {G}_{6,n},\vect {G}_{7,n}\right)
\label{eqn_recupera_leg_G}
\end{equation}
\begin{equation}
{\vect{g}}=
\left[
\begin{array}{ccccccc}
\vect{g}^T_{1,a}   & \vect{g}^T_{2,n}   &   \vect{g}^T_{3,n}  &   \vect{g}^T_{4,n} & \vect{g}^T_{5,n}   &   \vect{g}^T_{6,n}  &   \vect{g}^T_{7,n}
\end{array}
\right]^{T}
\label{eqn_recupera_leg_g}
\end{equation}
where $a$ denotes analytical resolution and, $n$ denotes numerical resolution of loop closures in the above equations. The submechanisms are denoted by $S_i$ where $i = 1 \dots 7$ as shown in Fig. \ref{fig:topological}. It is tested on the real system where the resolution of loop closures happens at 1kHz in real time including the calculation of inverse dynamics in actuation space.

\subsubsection{Overall Range of Motion}
%The Range of Motion (ROM) can provide a usable workspace for the leg joint movement when employed for rehabilitation exercises, as reported in~\cite{tijjani2022finding}. The spine and prismatic knee joints demonstrate linear movement in meters, whereas the hip and ankle mechanisms feature three distinct types of revolute movements involving rotations in radians around the $X$, $Y$, and $Z$ axes of the global coordinate frame. Table \ref{tab:rom} presents the ROM for all actuators in the system. %%The velocity limits are theoretical assertions, not verified throgh measurements. Therefore, the performance of the linear and rotary joints relies on the maximum values of their actuator pitch length, gear ratio, and power electronics to achieve optimal performance.
The study discussed in~\cite{Kumar2018design},~\cite{tijjani2022finding} explores the Range of Motion (ROM) capabilities of the Recupera-Reha lower extremity joints for improved workspace. %The exoskeleton consists of a variety of actuators. The spine joint consists of 6 DOF active joints and six independent joints: three prismatic joints translating along the X, Y, and Z axes and three revolute roll, pitch, and yaw joints. Similarly, the prismatic knee joint functions as the active and independent joint. Furthermore, three actuators each exists within the hip and ankle mechanisms, incorporating three distinct revolute joints operating as the independent joints.
The ROM, maximum force and torque, and maximum velocity available at each actuator joint of the exoskeleton full hybrid model is listed in Table \ref{tab:rom1}. Table \ref{tab:rom2} provides details of the independent joints, outlining the ROM used in the tree-abstraction model of the exoskeleton.
%and velocity limitations essential for the closed-loop mechanism of the exoskeleton, which forms a tree-abstraction model.

\begin{table}[!ht]
	\centering
	%\caption{Overview Technical Specification.}
	\caption{ROM for the Actuator Joints}
	\begin{tabular}{cccc}
		\toprule
		Actuators & Force$/$ & Max. Vel. & ROM \\
				 & Torque &  & \\
		\midrule
		6 Spine Actuators   &  570 N & 0.34 m/s & [\,0 , 0.11\,]\,m  \\
        Knee Actuator  & 662 N    &  0.34 m/s & [\,-0.064 , 0.09\,]\,m \\ %0.46\,m to 0.71\,m
		3 Hip Motors  &  176 Nm & 2.39 rad/s & [\,-0.436 , 0.436\,]\,rad \\
		3 Ankle Motors &  28 Nm & 7.17 rad/s & [\,-0.436 , 0.436\,]\,rad \\
		\bottomrule
	\end{tabular}
	\label{tab:rom1}
\end{table}

\begin{table}[!ht]
	\centering
	%\caption{Overview Technical Specification.}
	\caption{ROM for the Independent Joints}
	\begin{tabular}{rrr}
		\toprule
		Independent joints & ROM \\ %& Max. Velocity  \\
		\midrule
		Spine X & [\,-0.143 , 0.122\,]\,m \\%& 0.1 to 0.26 m/s
		Spine Y & [\,-0.153 , 0.153\,]\,m \\%& 0.1 to 0.26 m/s
		Spine Z & [\,-0.056 , 0.057\,]\,m \\%& 0.1 to 0.26 m/s
		Spine Roll & [\,-0.576 , 0.585\,]\,rad \\%& 0.1 to 0.26 m/s
		Spine Pitch & [\,-0.576 , 0.576\,]\,rad \\%& 0.1 to 0.26 m/s
		Spine Yaw & [\,-1.518 , 1.518\,]\,rad \\%& 0.1 to 0.26 m/s
        Knee joint  & [\,-0.064 , 0.09\,]\,m  \\%& 0.15 to 0.34 m/s \\ %0.46\,m to 0.71\,m
		Hip Roll  & [\,-0.349 , 0.646\,]\,rad \\
		Hip Pitch & [\,-0.262 , 0.611\,]\,rad \\
		Hip Yaw  & [\,-0.349 , 0.646\,]\,rad \\

		Ankle Roll & [\,-0.349 , 0.646\,]\,rad \\
        Ankle Pitch & [\,-0.262 , 0.611\,]\,rad \\
        Ankle Yaw & [\,-0.349 , 0.646\,]\,rad \\
		\bottomrule
	\end{tabular}
	\label{tab:rom2}
\end{table}

\subsection{Tree Abstraction Model}
\begin{figure}[!h]
	\centering
	\includegraphics[width= 0.48\textwidth]{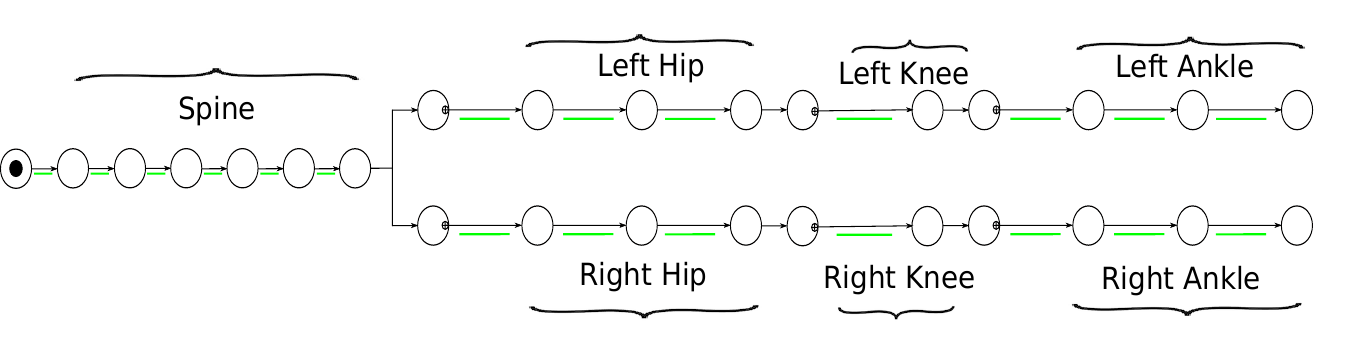}
	\caption{Tree-abstraction topology of the Recupera-Reha system}
	\label{fig:topological2}
	\vspace{-0.5cm}
\end{figure}

The full hybrid model in Fig. \ref{fig:topological} is complex in nature due to multiple loop closures . All the submechanisms in the system impose a total of 102 constraints ($n_c =102$) to be resolved. Although HyRoDyn can promptly resolve these constraints in real-time, integrating them into an optimization process poses a considerable challenge. Incorporating the intricate hybrid system into an OC framework necessitates derivatives for the loop closures, which are presently unavailable. Consequently, integrating these constraints into motion generation becomes inherently complex. As a result, a simplified tree-abstraction model of the entire system is typically preferred. The tree-abstraction model of the Recupera-Reha exoskeleton consisting of 20 DOF ($m = 20$) is shown in Fig. \ref{fig:topological2}, which can now be used for motion generation. %The tree abstraction model of the robot consist of 20 DOF ($m = 20$) that can be used for motion generation.
\vspace{-0.5cm}

\section{Motion Generation Defined As An Optimization Problem}
\label{sec_motion_generation}
In this section, we describe the motion generation for the Recupera-Reha exoskeleton using OC. This includes the constrained multibody dynamics up to the detailed formulation of the optimal control problem (OCP) for each movement.
% In this section, we define motion generation using the optimal control method. It describes from the constrained multi-body dynamics to the detailed formulation of the optimal control problem (OCP) of each movement.
From this section onwards, the tree-abstraction model will be considered, where $\q$ $\in \mathbf{R}^m$ is the vector of independent joint coordinates including the virtual floating-base joints.
\subsection{Whole-body dynamics under constraints}
\label{subsec:dyn}
The exoskeleton is a rigid, articulated multi-body system whose dynamics include $K$ contact constraints.
The equation of motion for this type of system is based on the Euler-Lagrange equations of motion (\ref{eq2}):
\begin{equation}
  \Mass(\q) \ddq + \Bias(\q,\dq) = \mathbf{S}^\intercal {\mathbf{\Torque}} + \sum_{k=1}^K {\mathbf{\Jac}}_{k}(\q)^\intercal {\mathbf{\Force_k}} 
  \label{eq2}
\end{equation}
where, 
\begin{itemize}
    \item $\q$ is the vector of joints' rotations and translations including the virtual floating-base joints.
    \item $\dq, \ddq$ are the vectors of generalized velocities and accelerations.
    \item $\Mass(\q)$ is the inertia matrix.
    \item $\Bias(\q,\dq)$ is the vector of gravity and non-linear forces.
    \item $\mathbf{S}$ is the selection matrix for the actuated elements.
    \item ${\mathbf{\Torque}} $ is the vector of internal joints torques.
    \item $\mathbf \Jac_{k}(\q)$ is the contact constraints' Jacobian matrix.
    \item ${\mathbf{\Force_{k}}}$ is the vector of contact forces.
\end{itemize}
%For further explanations, refer to~\cite{featherstone2014rigid}. 
Given that the system dynamics are defined in the acceleration dimension, the $k^{th}$ contact is expressed as a second-order static constraint (\ref{eq:cont_const}):
\begin{equation}
  \label{eq:cont_const}
  \mathbf \Jac_{k} \ddq + \dot{\mathbf{\Jac}_{k}} \dq = \mathbf 0 \hspace{1em} \forall k \in 1\cdots K
\end{equation}

The constrained dynamics is expressed as the combination of (\ref{eq2}) and (\ref{eq:cont_const}) in  (\ref{eq5}) :

\begin{equation}
  \label{eq5}
  \BIN
  \Mass & {\mathbf{\Jac}_{c}}^\intercal\\
  {\mathbf{\Jac}_{c}} & \mathbf 0
  \BOUT
  \BIN
  \ddq\\
  -\mathbf{\Force_{c}}
  \BOUT =
  \BIN
  \mathbf{S}^\intercal {\mathbf{\Torque}} - \Bias \\
  -\dot{\mathbf{\Jac}_{c}} \dq

  \BOUT
\end{equation}

$\Jac_{c} = \BIN \mathbf{\Jac_{1}}^\intercal \cdots \mathbf{\Jac_k}^\intercal \cdots \mathbf{\Jac_{K}}^\intercal\BOUT^\intercal$,  $\mathbf{\Force_{c}} = \BIN \mathbf{\Force_{1}}^\intercal \cdots \mathbf{\Force_k}^\intercal \cdots \mathbf{\Force_{K}}^\intercal\BOUT^\intercal$
\subsection{Optimal Control Problem Definition}
\label{subsec:ocp}
The trajectory optimization problem is descretized and formulated as follow:
\begin{subequations}
  \label{croc:eq4}
    \begin{eqnarray}
 \min_{\mathbf  x,\mathbf  u} \hspace{0.2em} l_N(\mathbf {x}_N) + \mathlarger{\sum}_{t=0}^{N-1} l(\mathbf {x}_t,\mathbf {u}_t) dt \\
    s.t. \quad \mathbf {x}_0 = \mathbf {f}_0, 
\\
  \forall i \in \lbrace 0 ... N-1 \rbrace, \hspace{0.5em} \mathbf {x}_{i+1} = \mathbf {f}_t(\mathbf {x}_i, \mathbf {u}_i)
  \end{eqnarray} 
\end{subequations}
where:
\begin{itemize}
	\item $\mathbf  x = (\q, \dq)$ is the state of the system,
	\item $\mathbf {u}$ is the torque control variable,
	\item $l_N$ is the Terminal cost model,
	\item $l$ is the Running cost model,
	\item $f_t$ is the discretization of the dynamics,
	\item $\mathbf{{x}_0}$ is the initial state of the system,
	\item $N$ is the number of nodes over a trajectory.
\end{itemize}
In this study, the OCP was resolved using a shooting method~\cite{Die17}, namely the Box-FDDP algorithm (DDP: Differential Dynamics Programming)~\cite{Bud18}.
The DDP algorithm is well suited to this type of optimization problem, as it exploits the system's sparsity in a high-iterative way.
FDDP (Feasibility DDP)~\cite{Man19} enables us to find a solution even from an infeasible guess, and overcomes the numerical restrictions of single-shot algorithms like the original DDP.
The box allows the resolution within the robot's permissible torque limits.
The open-source software \textit{Crocoddyl}~\cite{Mas20} with its Box-FDDP solver was employed for this resolution. 
The dynamics and their derivatives have been computed in a fast and efficient manner using the open-source software \textit{Pinocchio}~\cite{carpentier2019pinocchio} .
\subsection{Desired Motions Formulated As Cost Models}
\label{subsec:cost}
In order to evaluate the robot's capabilities, two movements have been designed as OCPs.
In the following, the OCPs will be described as cost models.
\subsubsection{Sitting and standing movements}
It's well known that sitting in a wheelchair for long periods of time leads to numerous health problems~\cite{li2016com}. %,liampas2021musc}.
The Recupera-Reha exoskeleton was originally designed for upper-body rehabilitation, making the ability to stand up and sit down an essential task for the system. This prevents the user from sitting for long periods.
This movement consists of a single phase for the sitting movement and a single phase for the standing movement.
The movement is designed such that the center of mass (CoM) is lowered for sitting and raised for standing, while maintaining the system's balance.
\subsubsection{Static walk movement}
The trajectory optimization allows to generate a wide variety of movements that add flexibility and versatility to the rehabilitation training.
With this in mind, a static walk has been designed while respecting the joints limits of the system.
This movement was selected to assess the robot's ability to move through its environment, in order to be able to provide lower limb rehabilitation exercises as well in the future.
The static walk movement represents only an example of the movements that could be designed or scaled to suit different walk models and rehabilitation needs.

The static walk movement is composed of a right and a left stride. Each stride is divided into two phases: the support phase and the swing phase.
\begin{itemize}
\item  \textit{Support Phase}: In this phase, the CoM (x,y) is shifted over the supporting foot before moving the opposite foot forward.
Both feet are in contact in this phase.
\item  \textit{Swing Phase}: In this phase, the swing foot breaks contact with the ground to move forward.
\end{itemize}
\subsubsection{Cost Models Definition}
To achieve these movements, the following Running and Terminal Cost Models have been implemented.
\begin{equation}
  \costl = {\mathlarger{\sum}_{c=1}^C} \alpha_c \Phi_c(\q,\dq,{\mathbf{\Torque)}}, \quad \costl_N = {\mathlarger{\sum}_{{c_F}=1}^{C_F}} \alpha_{c_F} \Phi_{c_F}(\q,\dq),
\end{equation}
Where, $\Phi_c , \Phi_{c_F}$ represent the cost of the running and terminal models respectively,  $ \alpha_c, \alpha_{c_F} \in \mathbb{R}$ represent their respective weights, which have been empirically determined.
In both movements, the cost functions composing the cost models are the following:
\begin{itemize}
  \item \textit{CoM target}: CoM placement $\mathbf{c}_w$ tracks the CoM reference placement at the end of each phase of movement.
    $\Phi_{1} = \parallel \mathbf{c}_w(t)-\mathbf{c}_w^\text{ref}(t_N)\parallel^{2}_2 \alpha_1$       
  \item \textit{Torque regularization}: Joint torques minimization to ensure the dynamic feasibility.\\
    $\Phi_{2} = \parallel {\mathbf{\Torque}} (t) \parallel^{2}_2 \alpha_2$
  \item \textit{Posture regularization}: This cost manages the redundancy of the multi-body dynamics. \\
    $\Phi_{3} = \parallel \q(t)-\q^{ref}(t_N)\parallel^{2}_2 \alpha_3$
\end{itemize}
$t_N$ reprsents the final time of each phase of the motion.

The static walk movement requires an additional cost, the foot-tracking cost, in order to move the swing foot forward.
\begin{itemize}
	  \item \textit{Foot Tracking}: The foot placement $\mathbf{r}_w$ tracks the final foot position of each phase (right or left, depending on the movement phase).\\
    $\Phi_{4} = \parallel \mathbf{r}_w(t)-\mathbf{r}_w^\text{ref}(t_N)\parallel^{2}_2 \alpha_4$
\end{itemize}
The terminal cost model excludes the torque regularization cost function because it prioritizes achieving the final state, while torque regularization is more relevant for optimizing the trajectory throughout the entire movement.

\section{Results}
\label{sec_experiment_and_results}
This section encapsulates the experimental pipeline and presents simulation and experimental results for the motions expressed as OCP problems (sitting, standing and static walking). The results are available in the accompanying
video\footnote{\url{https://youtu.be/mL4wIBC-q3s}}

\subsection{Experimental pipeline}
\label{subsec:8}
The experimental pipeline in this work utilizes the specific movements generated from OC method within the independent joint space. Additionally, we employ the HyRoDyn software framework to calculate actuation space trajectories for testing on the real system.
To this end, we first formulate these movements as constrained OCP using CROCODDYL software.
We validate the resulting trajectories in simulation, providing graphical results in the form of plots to assess the dynamic feasibility of these movements. The HyRoDyn framework then transmits these validated movements to the real robot, mapping the motions of the independent joints into the actuation joint space. To evaluate the executed movements, we use a graphical plot analysis for joint tracking.

\subsection{Results from simulation and experiment}
\label{subsec:18}

\subsubsection{Sitting and standing motion}
\label{subsec:9}
\begin{figure}[!ht]
\vspace{-0.2cm}
\centering
\subfloat[\centering Simulation]{\includegraphics[width= 0.25\textwidth]{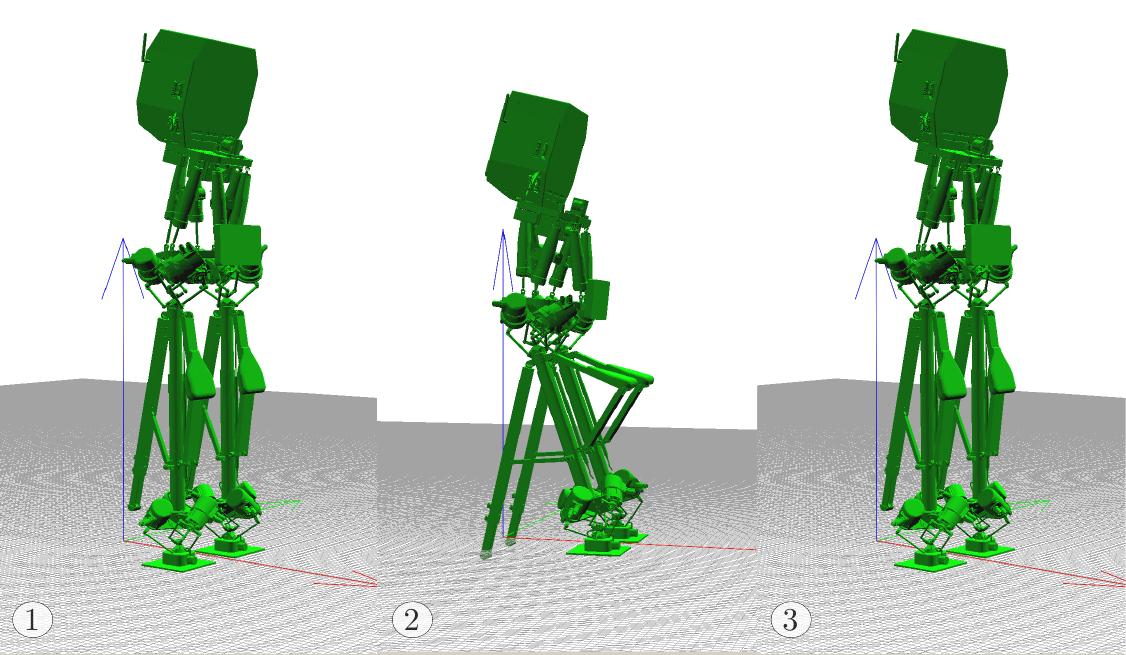}}
\hfill
\centering
\subfloat[\centering Real system]{\includegraphics[width= 0.25\textwidth]{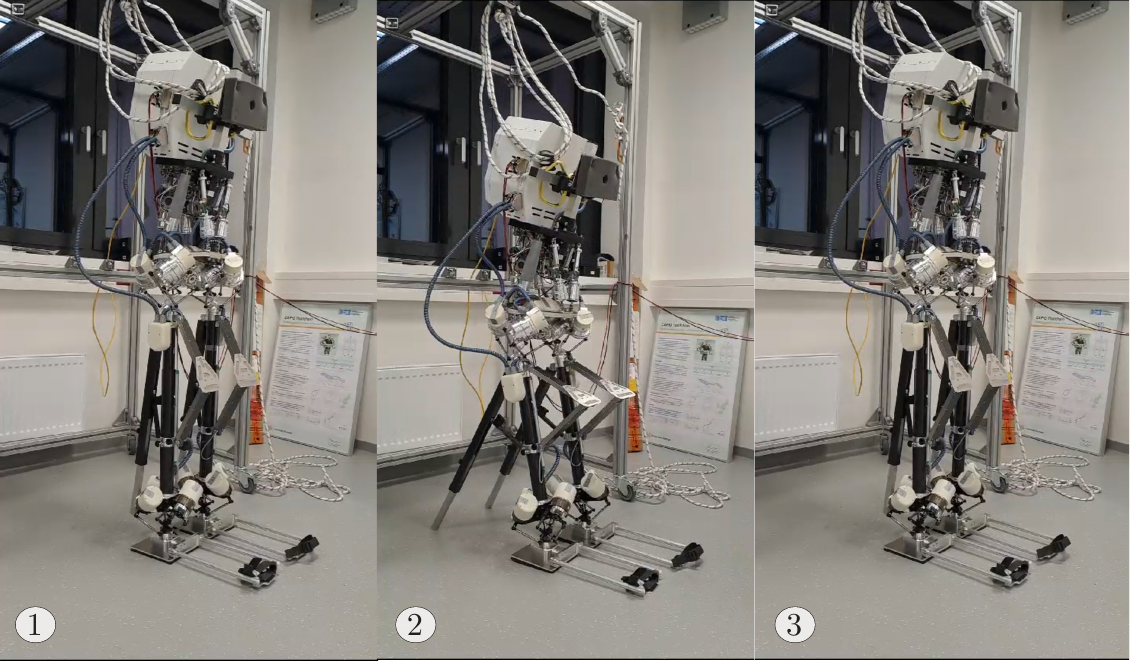}}
\caption{The screenshot of the exoskeleton at initial standing, sitting, and final standing position, both in simulation and on the real system}
\label{fig:sitandstand}
\vspace{-0.6cm}
\end{figure}

Fig.~\ref{fig:sitandstand}(a) illustrates the initial standing, sitting, and final standing postures for the simulated sitting and standing motions. Similarly,
Fig.~\ref{fig:sitandstand}(b) depicts the experimental illustration of these motions.

\begin{figure}[!htb]%[H]%
	\centering
	\subfloat[\centering Sitting]{{\includegraphics[scale=0.17]{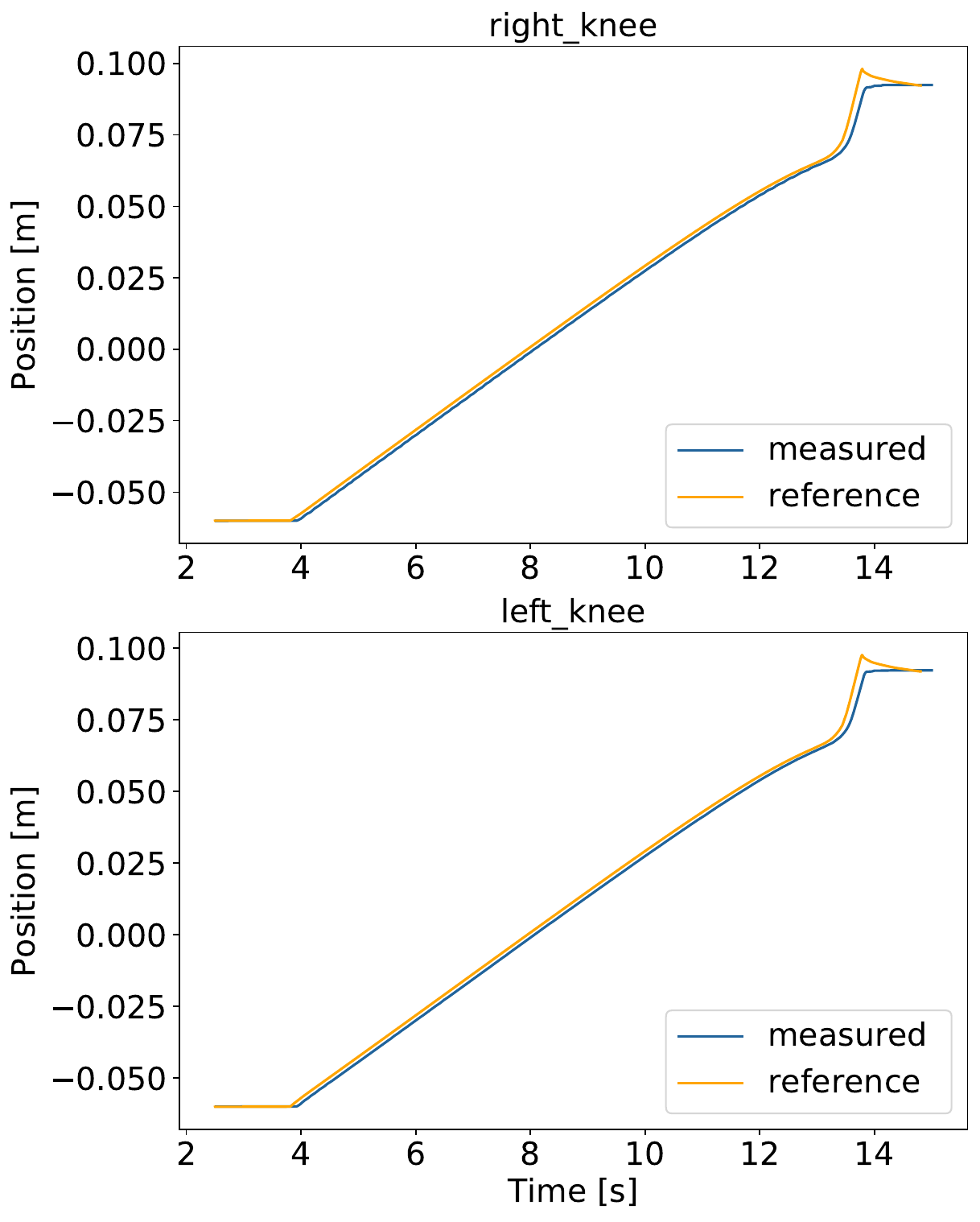} }\label{fig:stand_sit-sitting}}
	%\qquad
	\subfloat[\centering Standing]{{\includegraphics[scale=0.17]{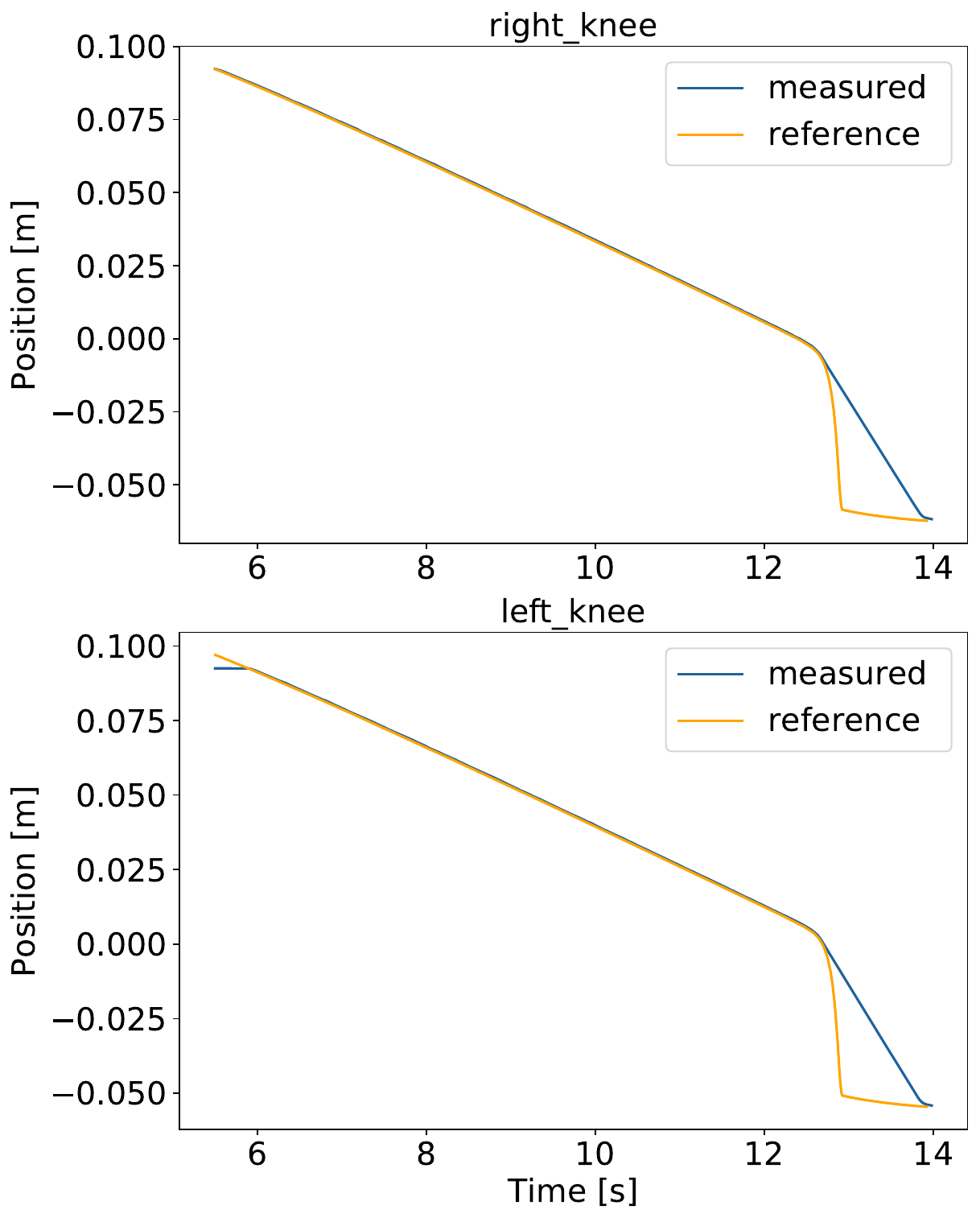} }\label{fig:stand_sit-standing}}
	\vspace{-0.1cm}
	\caption{Knee joint actuators in sitting and standing motion}
	\label{fig:stand_sit}
	\vspace{-0.4cm}
\end{figure}

\begin{figure}[!ht]%[!ht]
	\centering
	\subfloat[\centering Right hip joint actuators]{\includegraphics[scale=0.18]{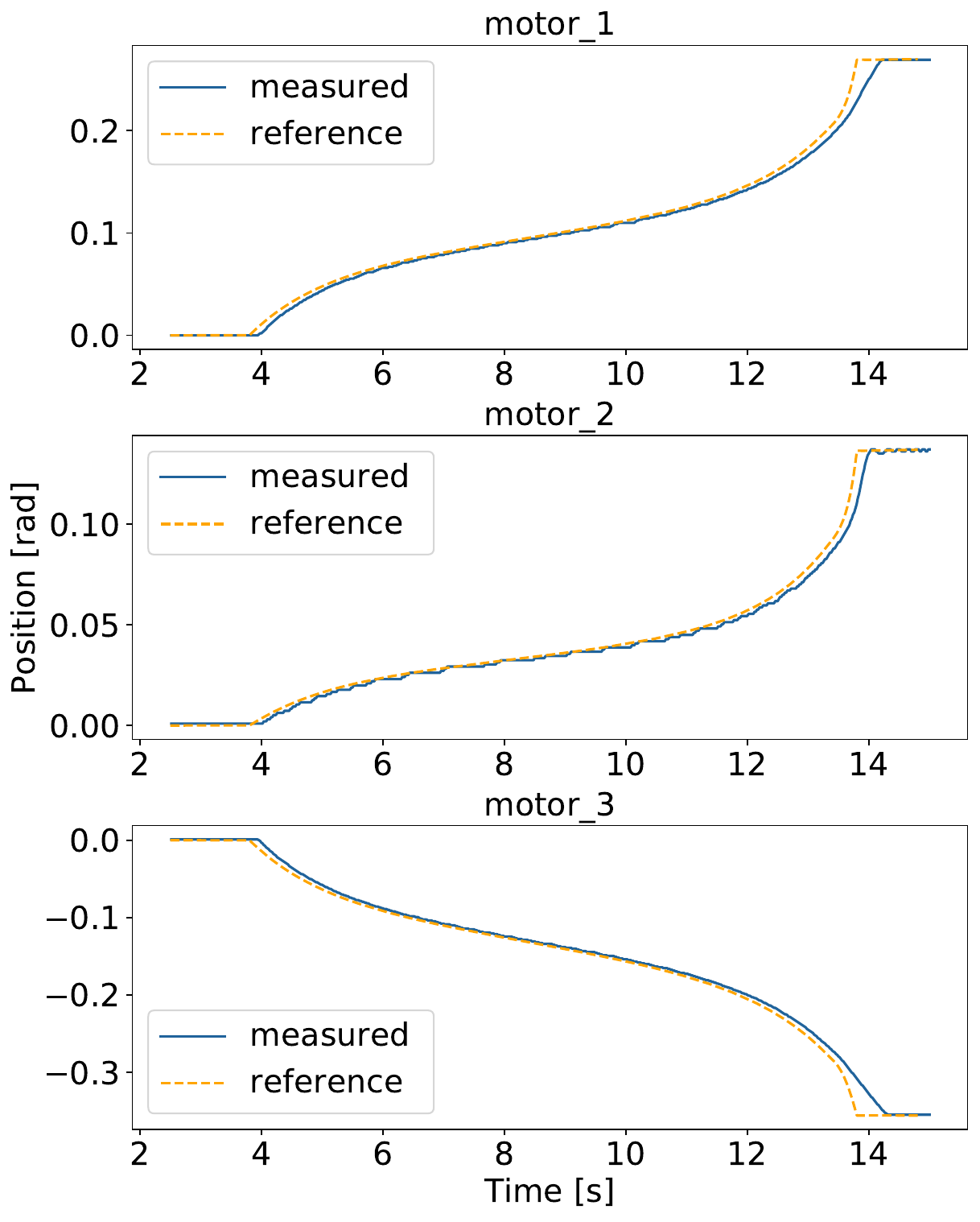}\label{fig:sitandstand-hip_sit}}
	%\hfill
	\centering
	\subfloat[\centering Left hip joint actuators]{\includegraphics[scale=0.18]{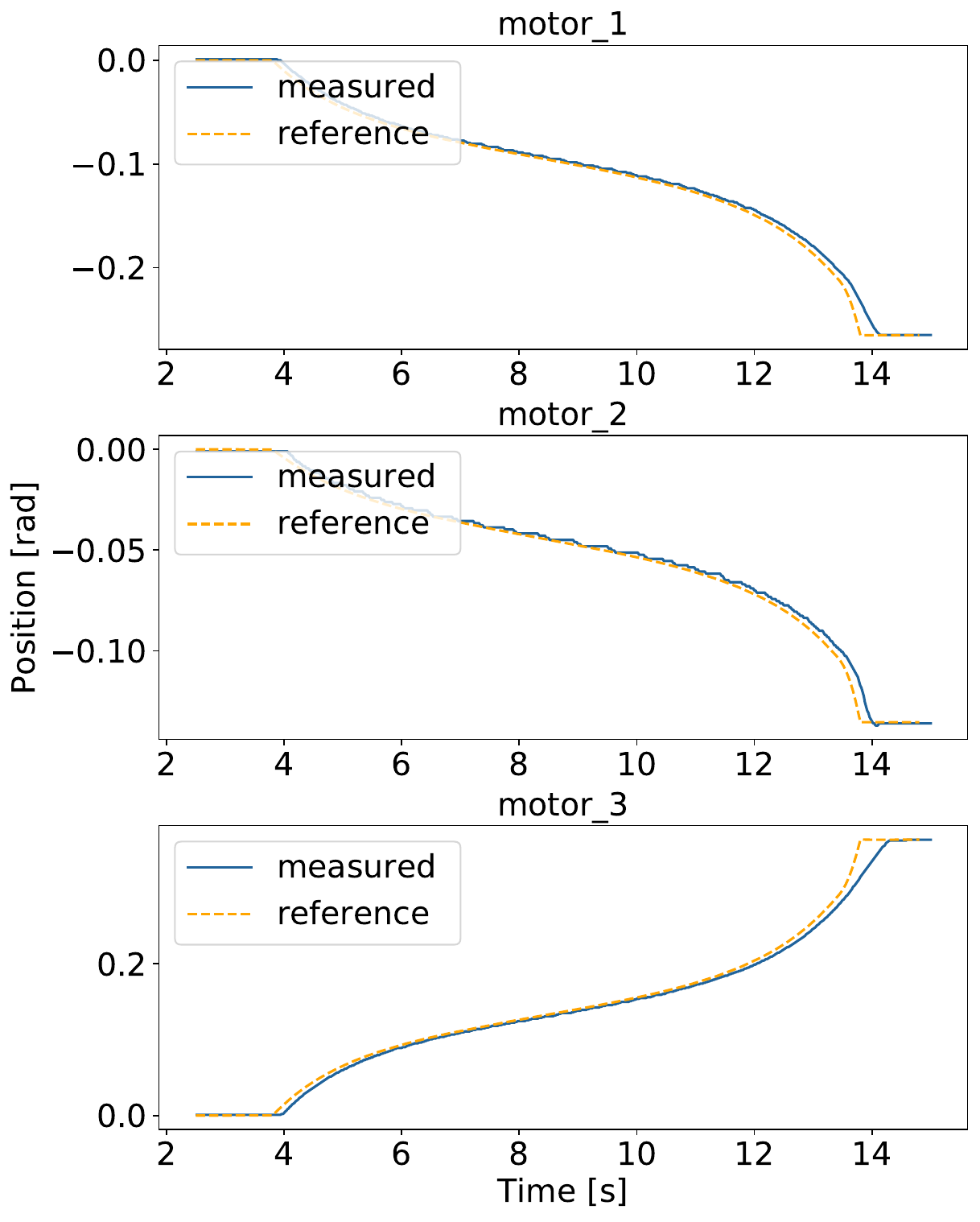}\label{fig:sitandstand-hip_stand}}
	\vspace{-0.1cm}
	\caption{Hip joint actuators during the sitting motion}
	\vspace{-0.4cm}
	\label{fig:sitting-hip}
\end{figure}

\begin{figure}[!ht]%[!ht]
	\centering
	\subfloat[\centering Right hip joint actuators]{\includegraphics[scale=0.18]{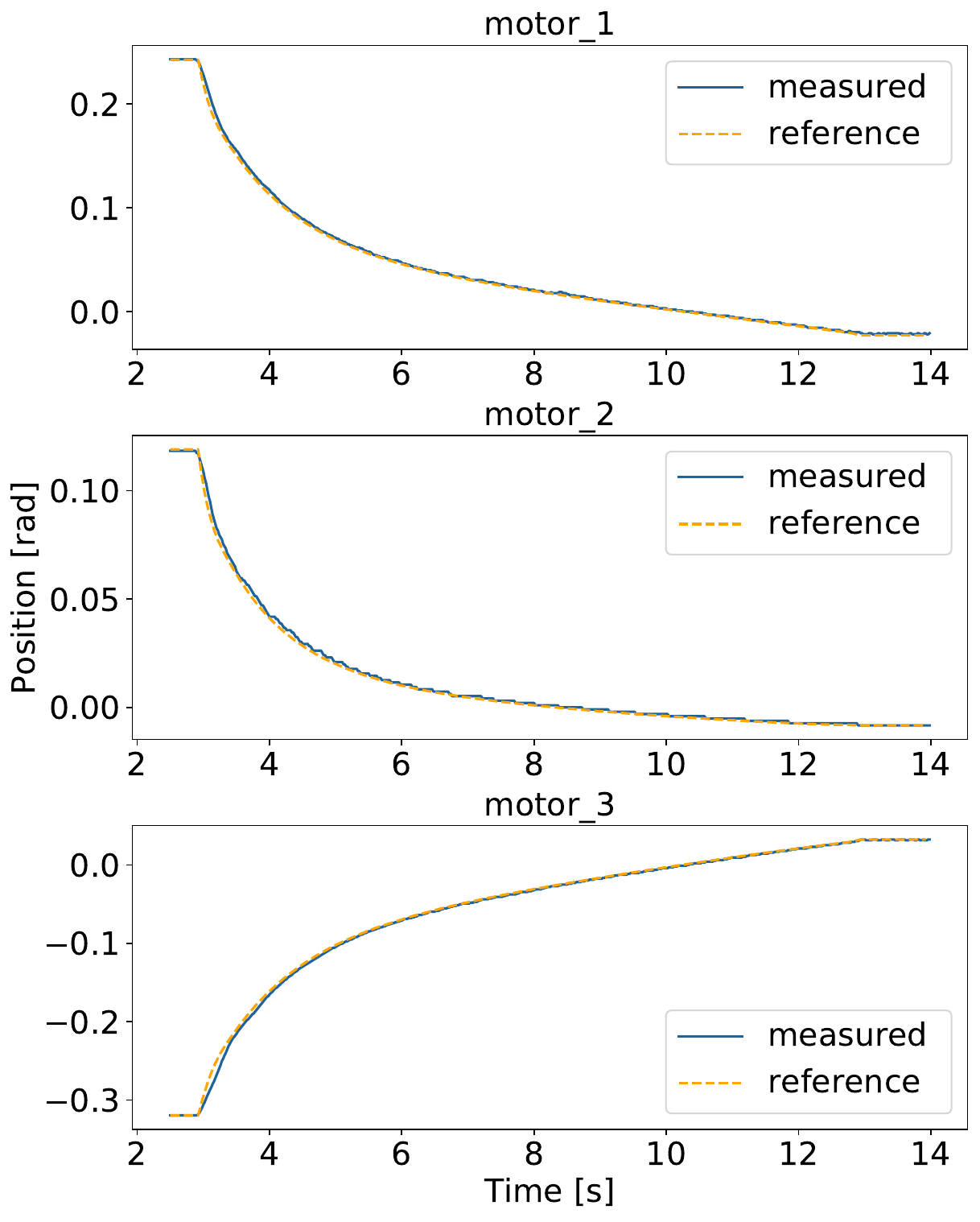}\label{fig:sitandstand-hip_sit}}
	%\hfill
	\centering
	\subfloat[\centering Left hip joint actuators]{\includegraphics[scale=0.18]{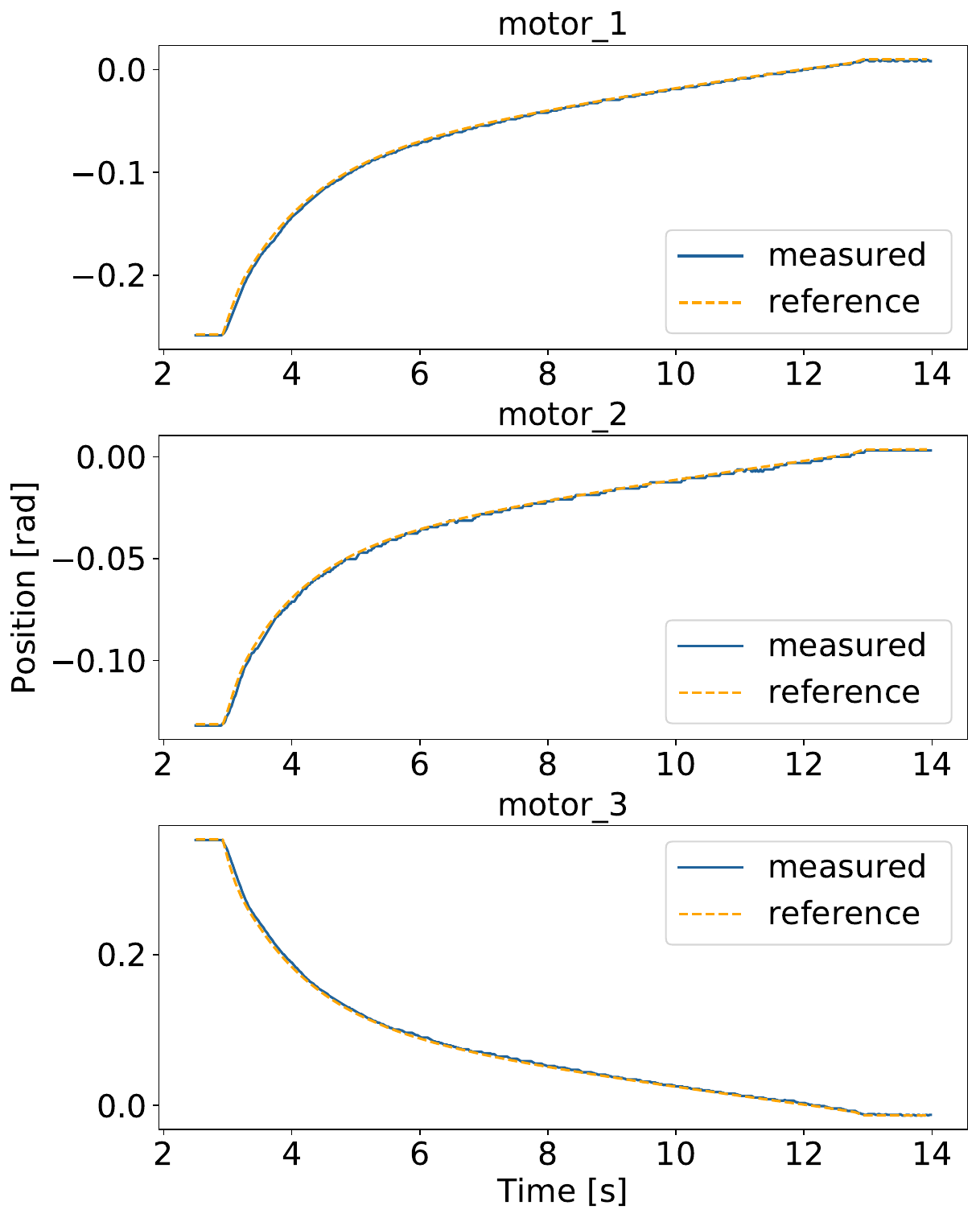}\label{fig:sitandstand-hip_stand}}
	%\vspace{-0.2cm}
	\caption{Hip joint actuators during the standing motion}
	\label{fig:standing-hip}
	\vspace{-0.4cm}
\end{figure}

For the sitting and standing motions, we have selected the prismatic knee and the hip joints as the primary joints involved in the motion.
The curves in Fig.~\ref{fig:stand_sit} show the movement of the right and left prismatic knee joints during the two transition motions. Similarly, Fig.~\ref{fig:sitting-hip} and~\ref{fig:standing-hip} depict the motions of the three actuators on each of the right and left hip joints during the motions, respectively. We can observe that the measured positions of the actuators aligned closely with the reference positions obtained from the OC method.
This alignment is evident in the smooth trajectory curves shown in the figures and confirmed by the video recordings of the motions. The consistency between the reference and measured trajectories demonstrates the robot's capability to execute the desired motions accurately. However, a deviation in the knee actuator position during standing was noted due to the higher energy needed to lift the robot against gravity. Despite this, the exoskeleton reached its target position.

\vspace{-0.4cm}

\vspace{-0.5cm}

\subsubsection{Static walk motion without a human}
\label{subsec:10}
\begin{figure}[!htbp]
\vspace{-0.1cm}
	%\hfill
	\centering
	\subfloat[\centering Walking in simulation]{\includegraphics[width= 0.28\textwidth]{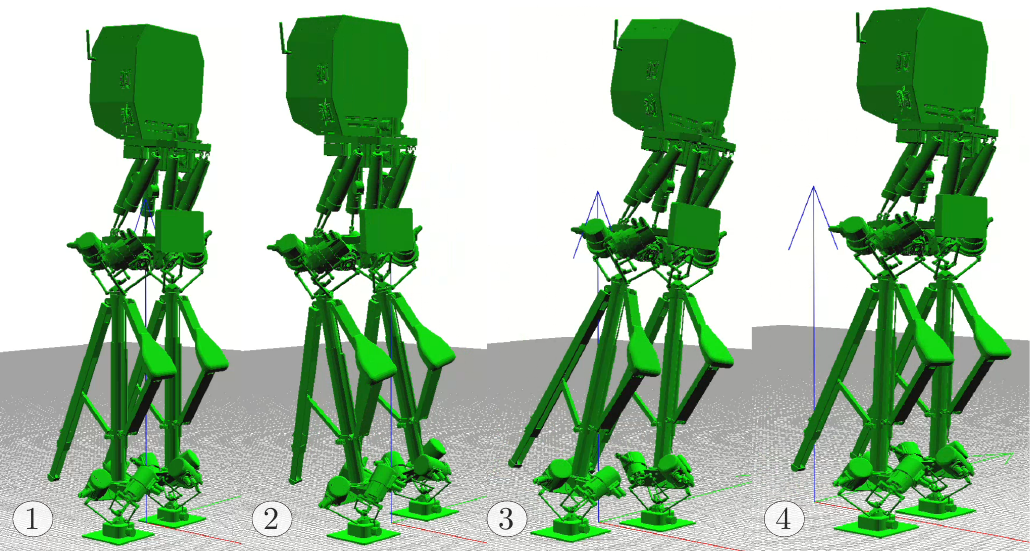}}
	\hfill
	\centering
	\subfloat[\centering Walking on real system]{\includegraphics[width= 0.28\textwidth]{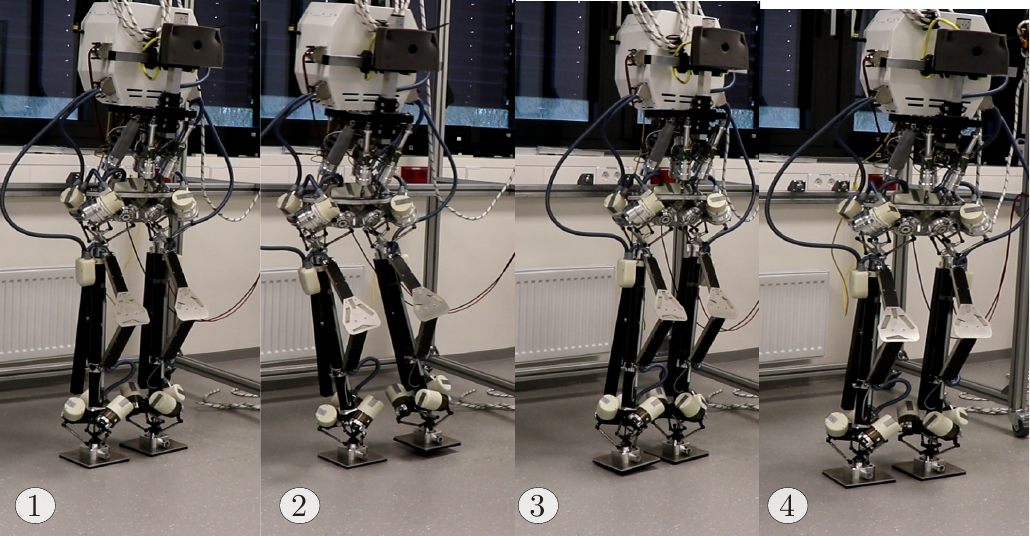}}
	\caption{The Recupera exoskeleton four-steps static walking motion in simulation and on the real system}
	\label{fig:walking_2}
	\vspace{-0.4cm}
\end{figure}

\begin{figure}[!htb]%[thpb]
\vspace{-0.2cm}
\centering
\subfloat[Right Knee joint actuator]{\includegraphics[width=4.0cm, height=2.9cm]{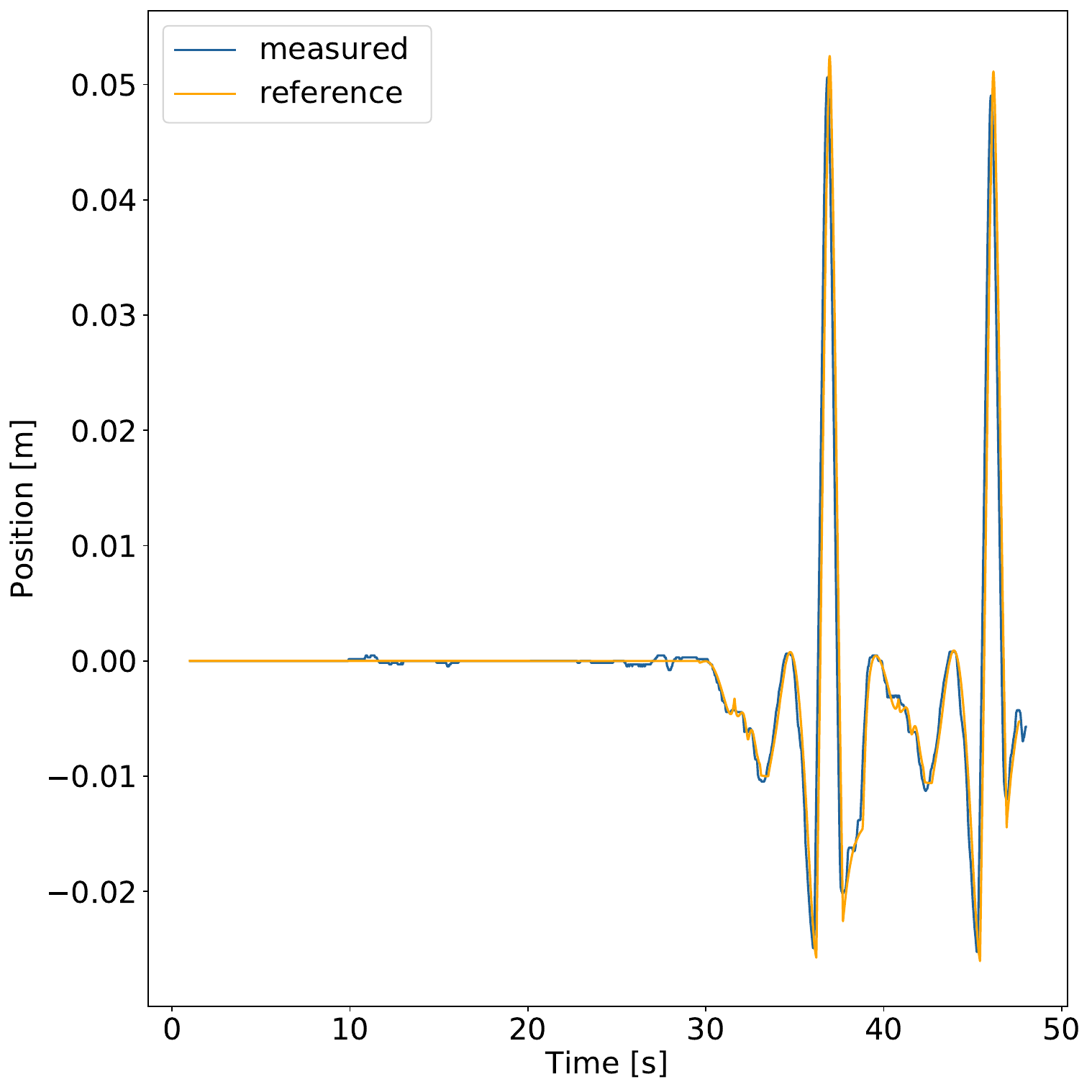}}
%\hfill
\subfloat[Left knee joint actuator]{\includegraphics[width=4.0cm, height=2.9cm]{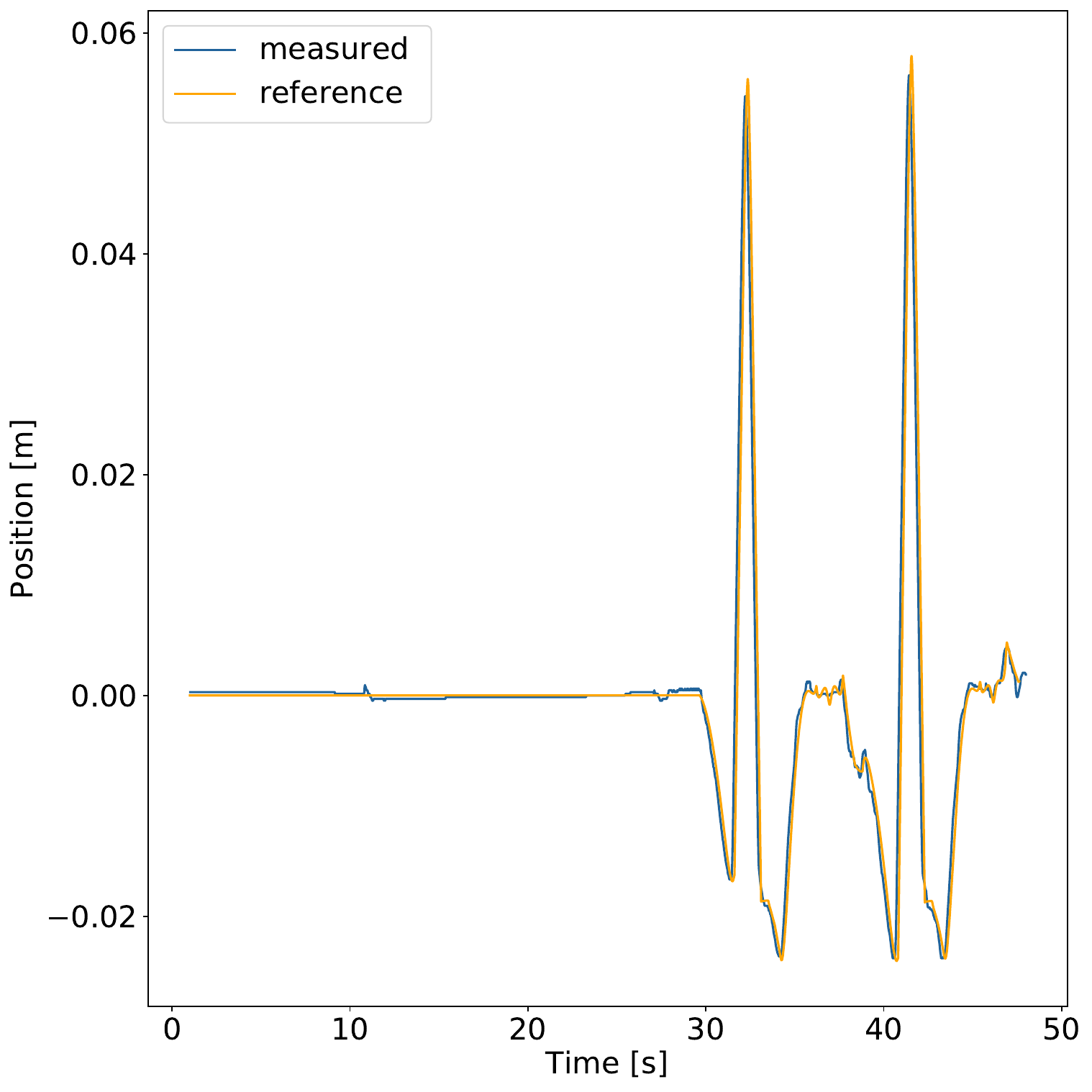}}
\vspace{-0.1cm}
\caption{Prismatic knee joint's actuator tracking}
\label{fig:walking-knee}
\vspace{-0.4cm}
\end{figure}

\begin{figure}[!htbp]
\centering
\subfloat[Right hip joint actuators]{\includegraphics[width=4.0cm]{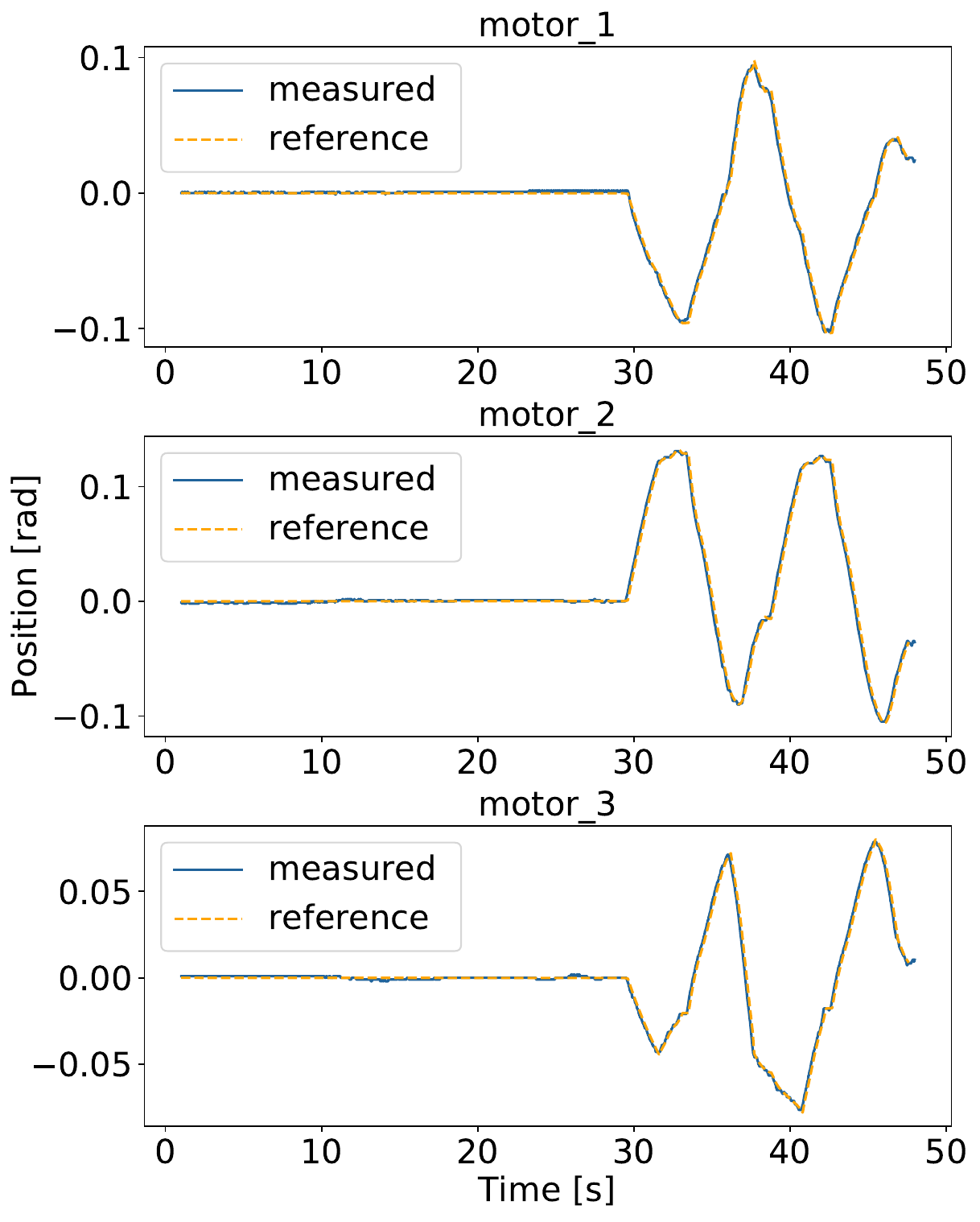}}
%\hfill
\subfloat[Left hip joint actuators]{\includegraphics[width=4.0cm]{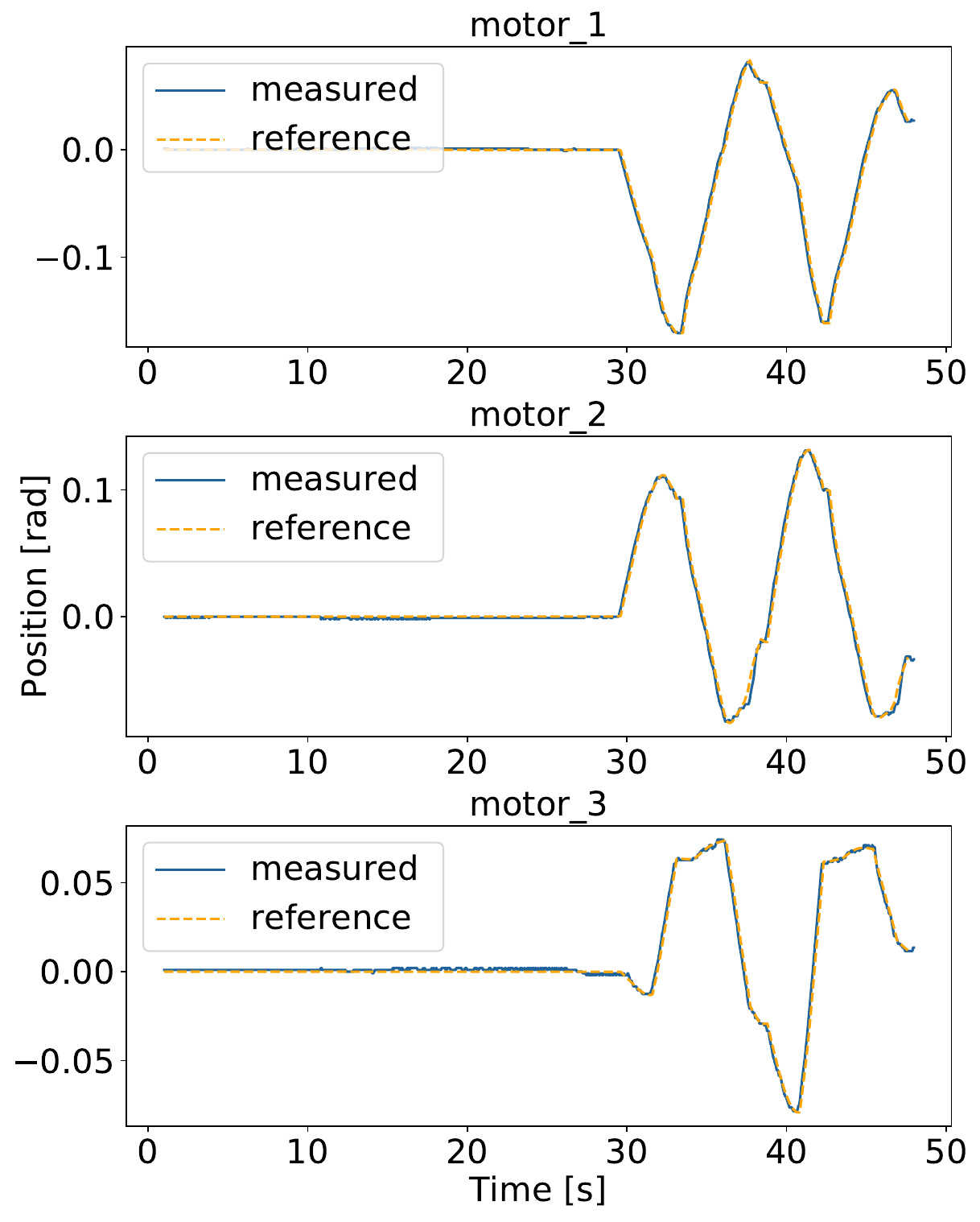}}
\caption{Hip joint's actuator tracking}
\label{fig:walking}
\vspace{-0.4cm}
\end{figure}

\begin{figure}[!htb]%[thpb]
\subfloat[Right ankle joint actuators]{\includegraphics[width=4.0cm]{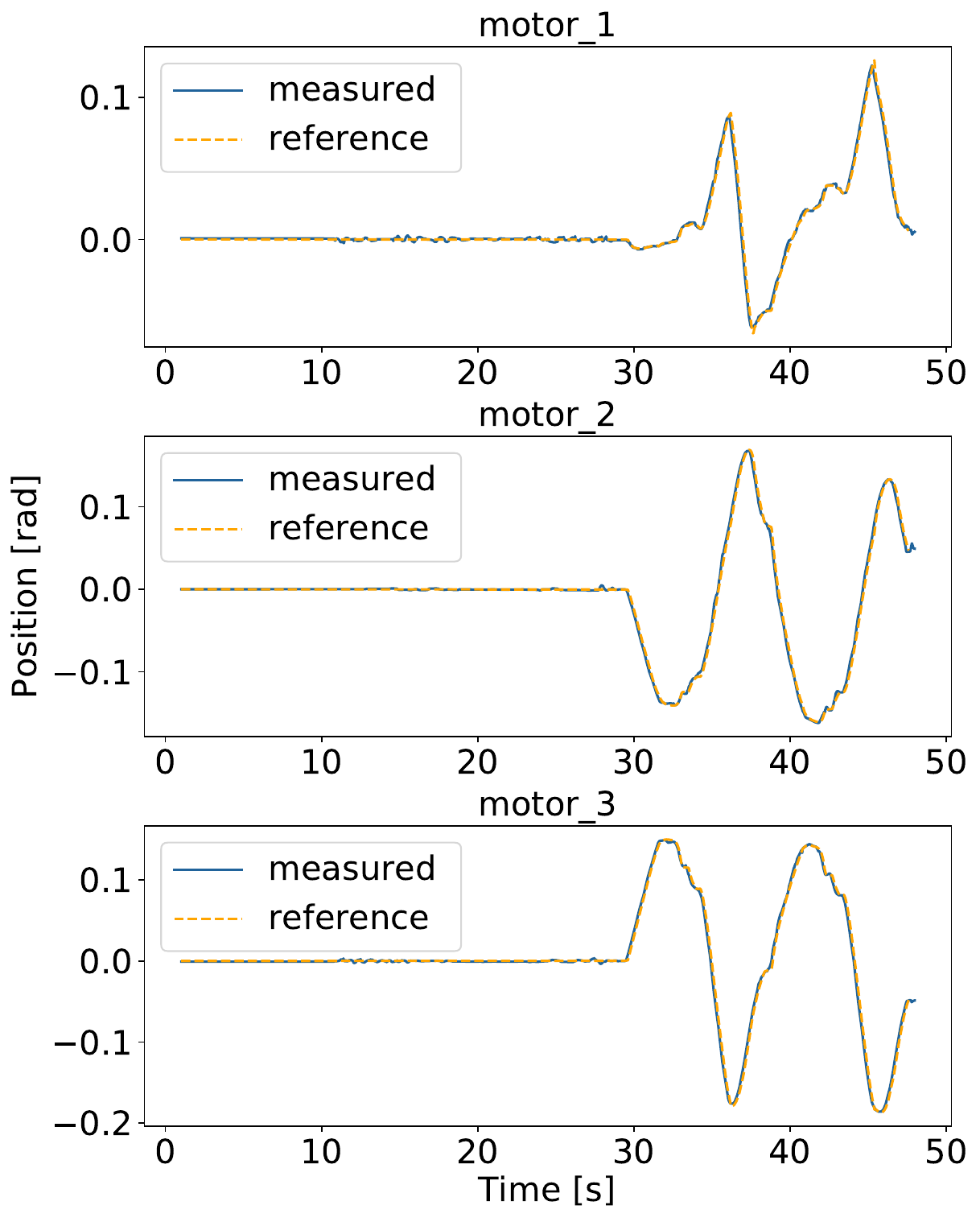}}
%\hfill
\subfloat[Left ankle joint actuators]{\includegraphics[width=4.0cm]{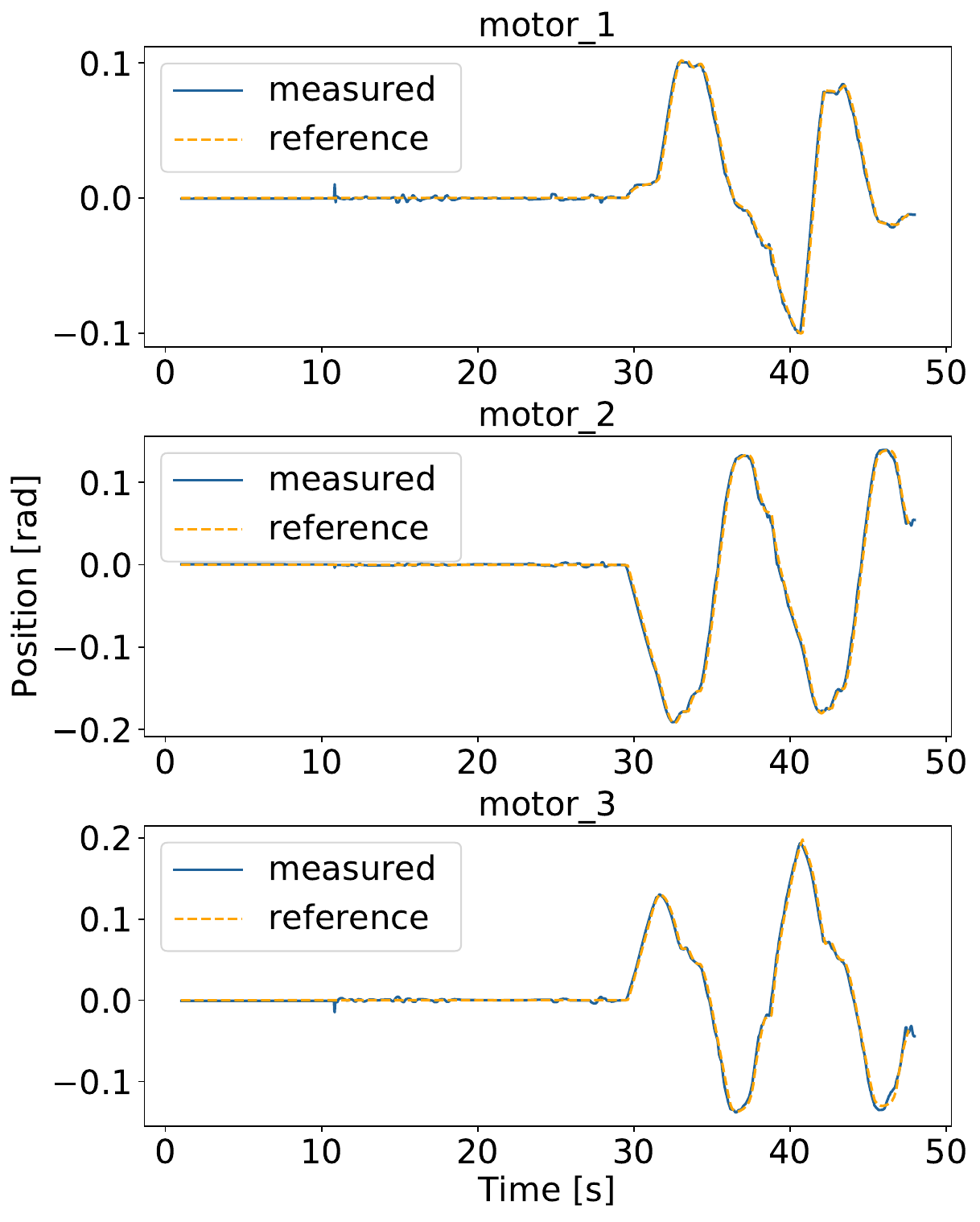}}
\caption{Ankle joint's actuator tracking}
\label{fig:walking-ankle}
\vspace{-0.4cm}
\end{figure}

We present the first results of exoskeleton walking motion without a human wearer. Fig.~\ref{fig:walking_2}(a) and Fig.~\ref{fig:walking_2}(b)
illustrates the following walking phases: 1) both legs at initial stance position; 2) left leg stride; 3) right leg stride; and 4) both legs at final stance position for the four-step static walking motion in simulation and on the real system. Figures~\ref{fig:walking},~\ref{fig:walking-knee}, and~\ref{fig:walking-ankle} show the actuator joint tracking, comparing the reference motion generated by OC with the measured actuator positions of the lower extremity joints during the experiment.
The close alignment between measured and reference position curves suggests a successful execution of the static walk motion on the real robot.

%\vspace{-0.4cm}
\subsubsection{Experiments with a human wearing the robot}
\label{subsec:11}
\begin{figure}[!ht]

%\hfill
\centering
\subfloat{\includegraphics[width= 0.45\textwidth]{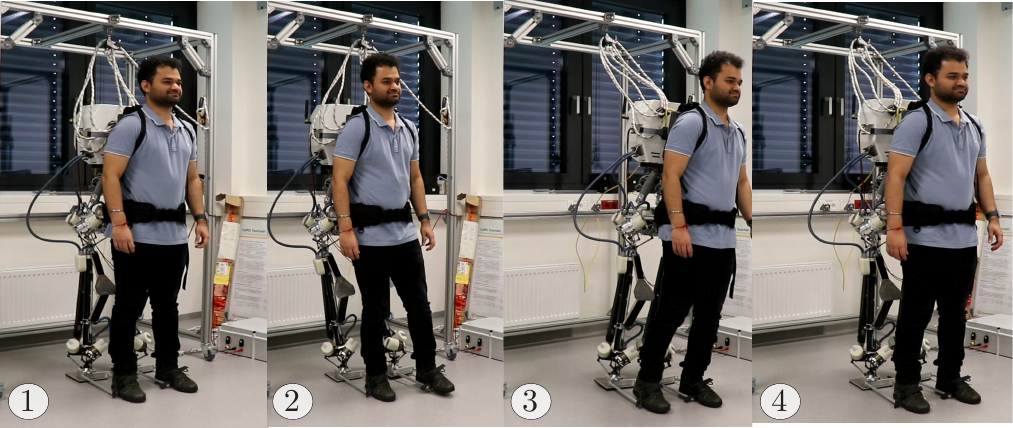}}
\caption{The screenshots of a human wearing the Recupera exoskeleton during the static walk experiment}
\label{fig:exo_with-human}
\vspace{-0.3cm}
\end{figure}

\begin{figure}[!htb]%[!ht]%[thpb]
\centering
\subfloat[Right hip joint actuators]{\includegraphics[width=4.cm]{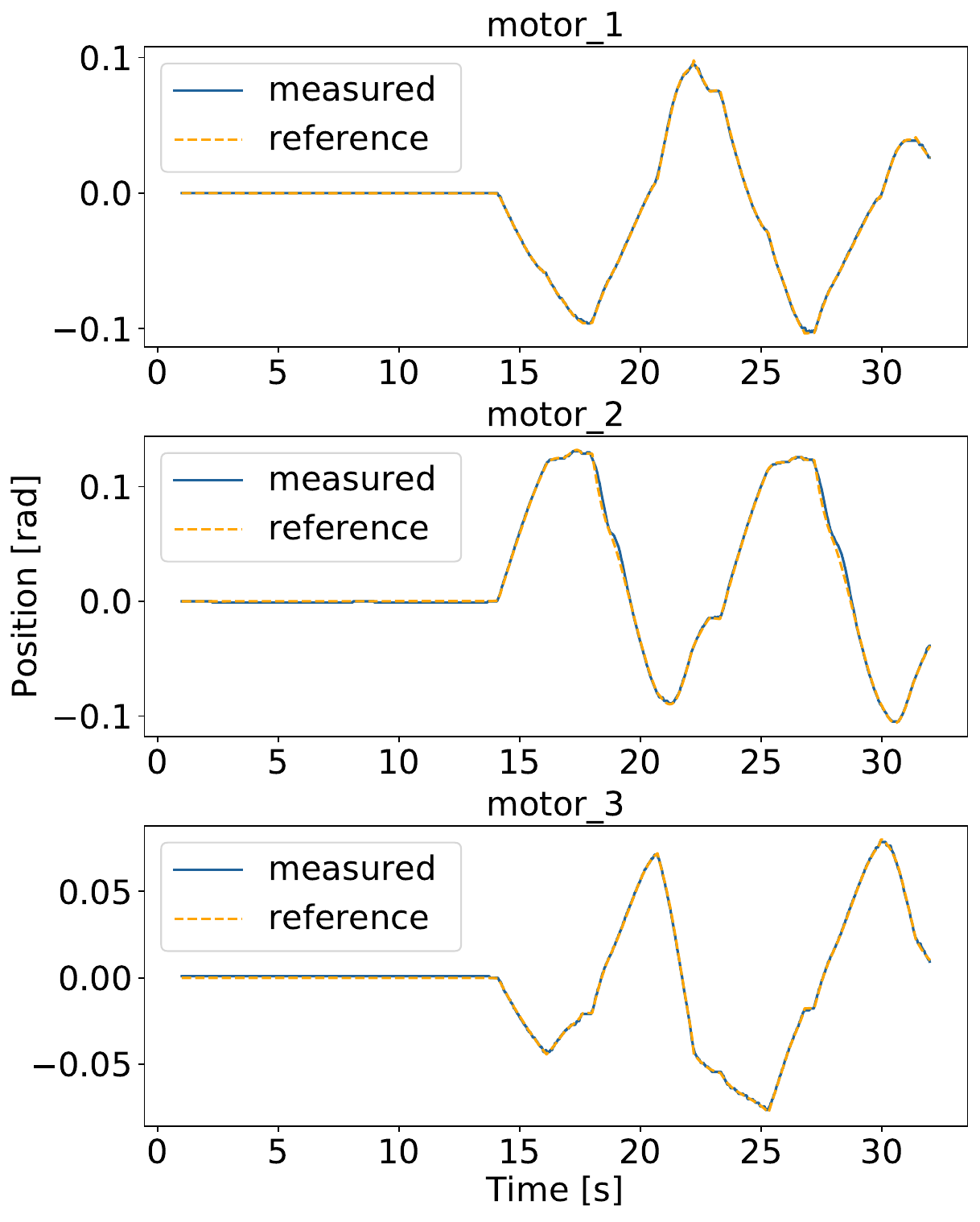}}
%\hfill
\subfloat[Left hip joint actuators]{\includegraphics[width=4.cm]{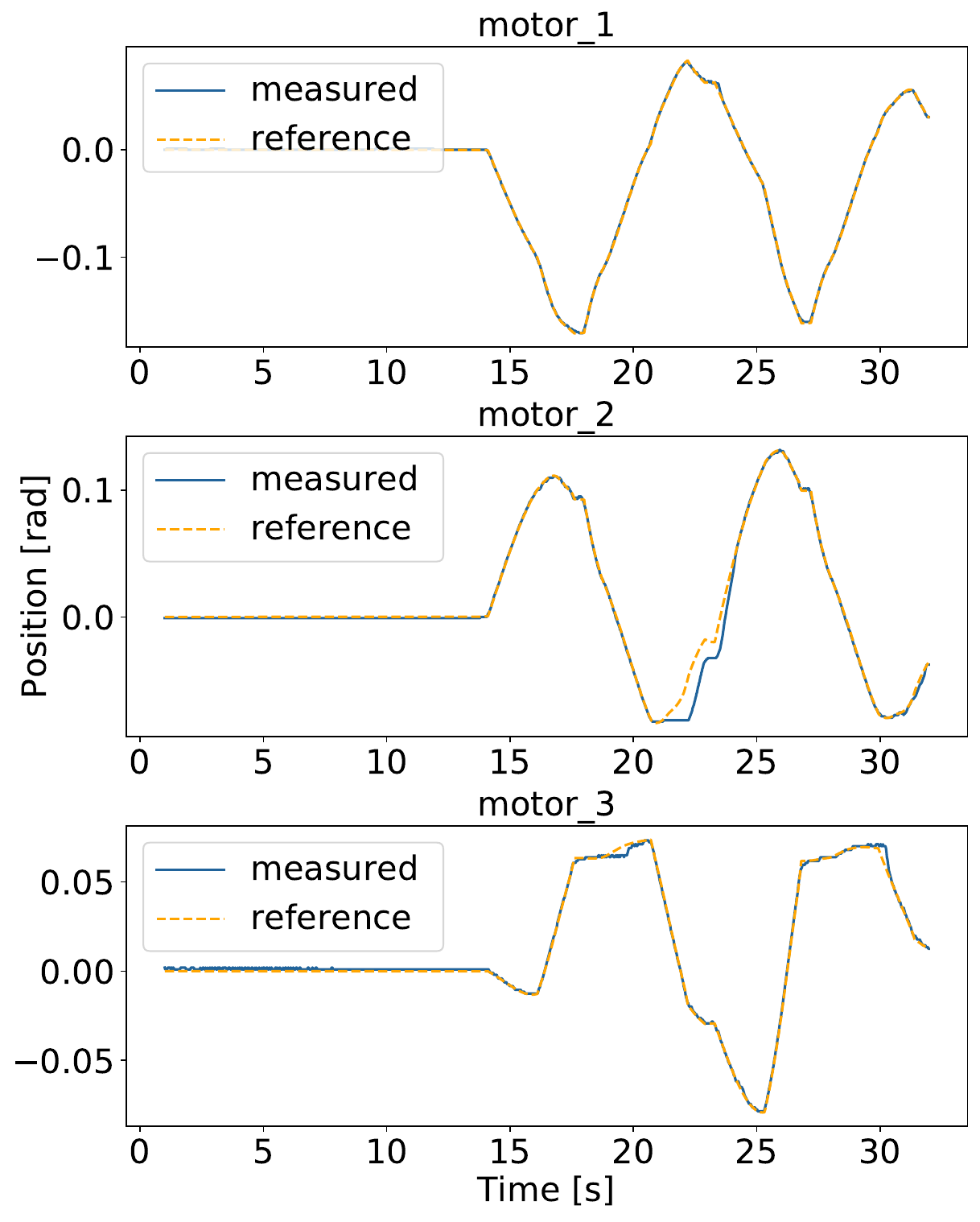}}
\caption{Hip joint's actuator tracking }
\label{fig:walkinghuman}
\vspace{-0.4cm}
\end{figure}

\begin{figure}[!htb]%[H]%[thpb]
\centering
\subfloat[Right knee joint actuator]{\includegraphics[width=4.0cm, height=3.0cm]{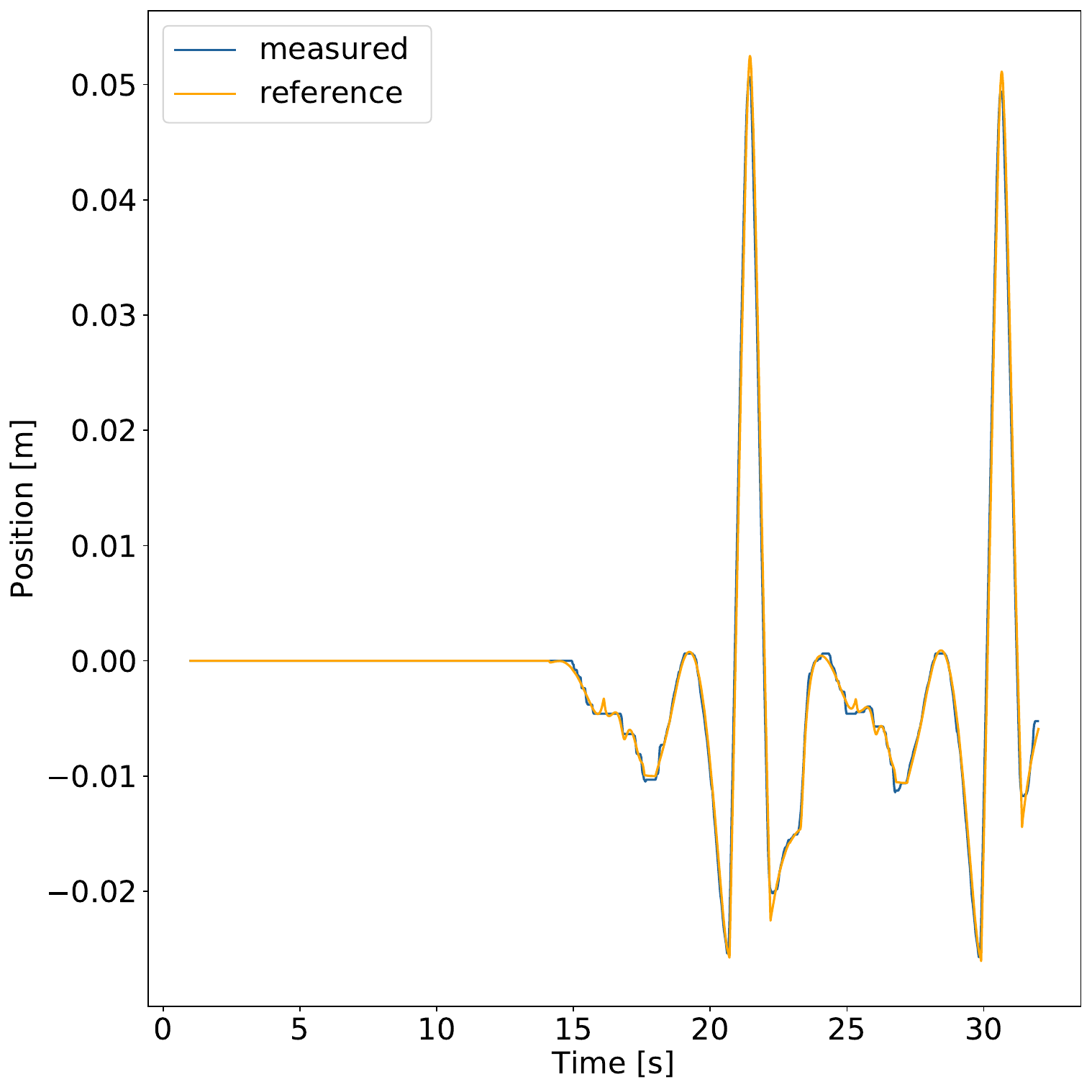}}
%\hfill
\subfloat[Left knee joint actuator]{\includegraphics[width=4.0cm, height=3.0cm]{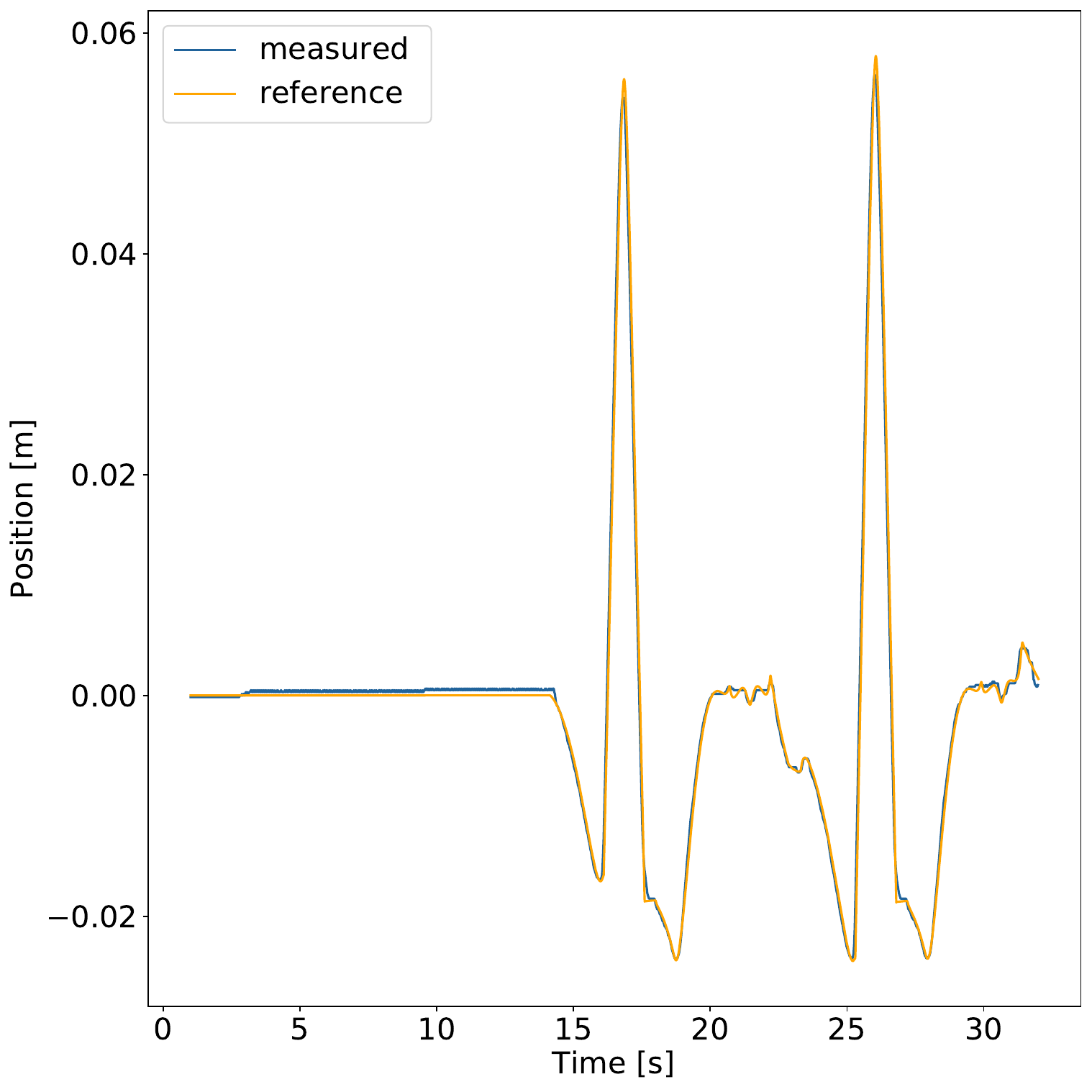}}
\caption{Prismatic knee joint's actuator tracking}
\label{fig:walking-kneehuman}
\vspace{-0.4cm}
\end{figure}

\begin{figure}[!htb]%[H]%[thpb]
\vspace{-0.1cm}
\subfloat[Right ankle joint actuators]{\includegraphics[width=4.1cm]{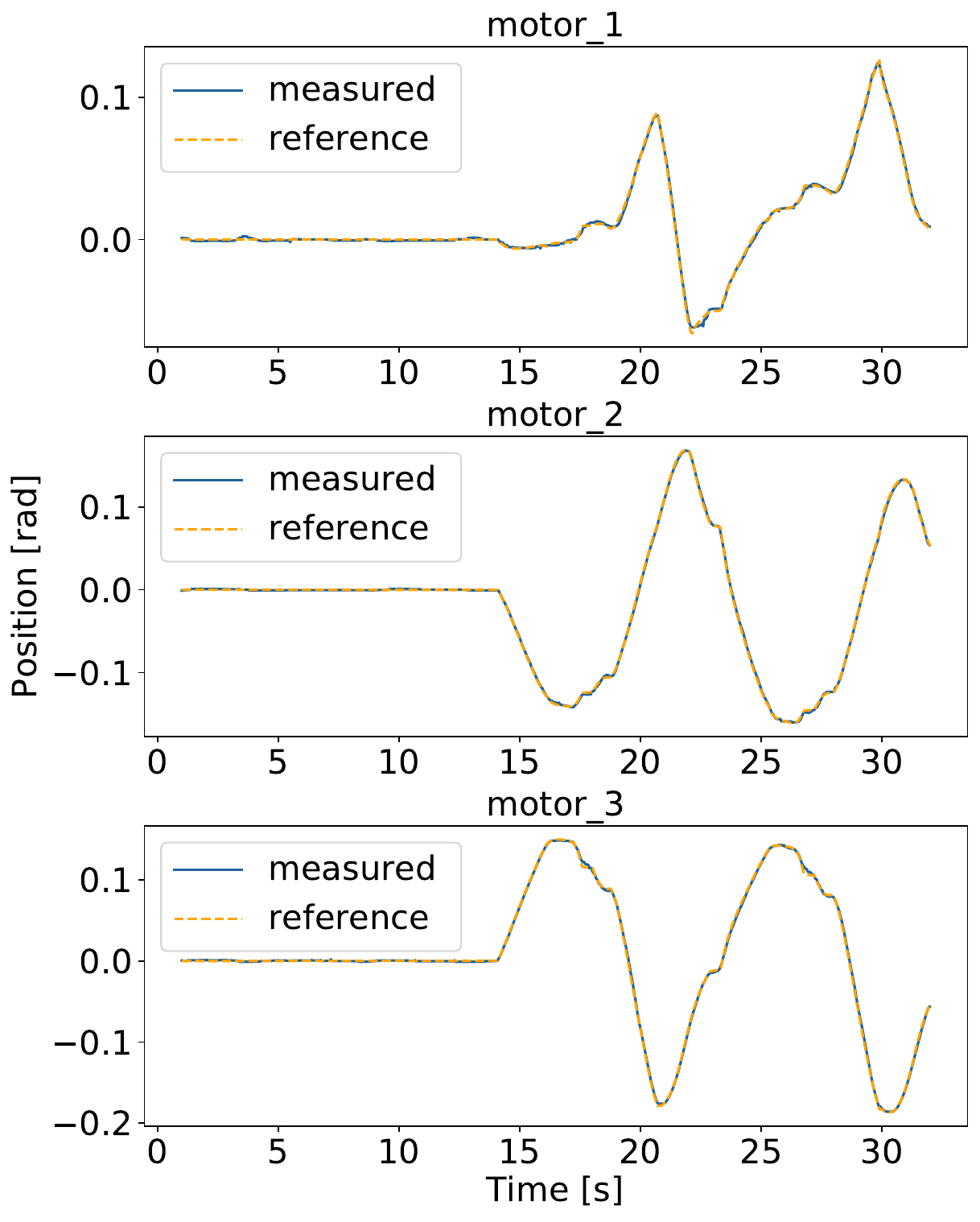}}
%\hfill
\subfloat[Left ankle joint actuators]{\includegraphics[width=4.1cm]{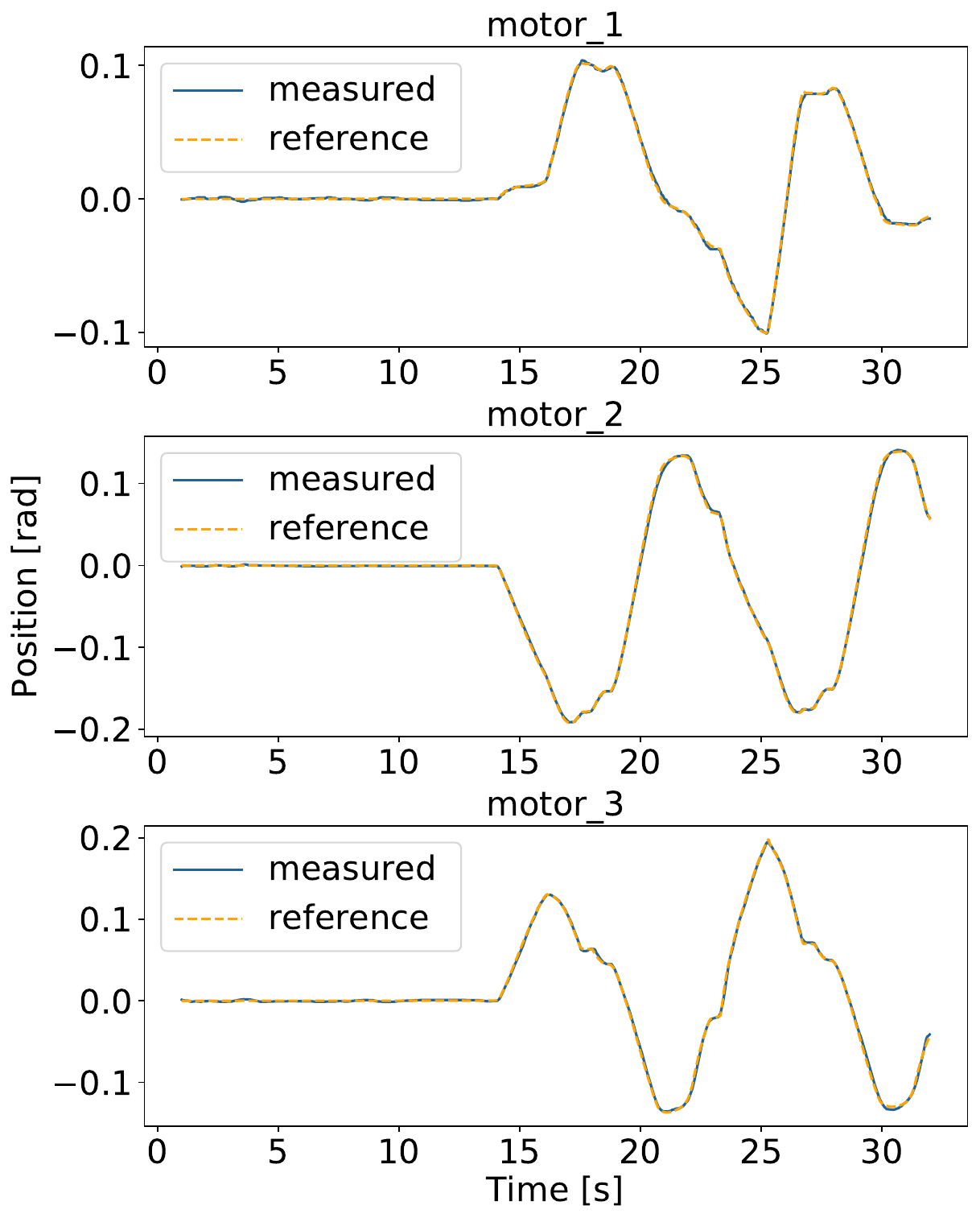}}
\caption{Ankle joint's actuator tracking}
\label{fig:walking-anklehuman}
%\vspace{-0.9cm}
\end{figure}

The screenshot in Fig.~\ref{fig:exo_with-human} shows a human wearing the exoskeleton during the four-steps static walking. The figure depicts the following walking phases: 1) both legs at stance position, 2) left leg stride, 3) right leg stride, and 4) both leg stance position for the fourth step.
The graphical representations of the four-step static walk with a human wearing the exoskeleton are depicted in Figures~\ref{fig:walkinghuman},~\ref{fig:walking-kneehuman}, and~\ref{fig:walking-anklehuman}, as well as in the accompanying video. Despite the added weight of the human, the joint actuators effectively compensated for the additional torque and accommodated the external force from the human wearer. We can observe a smooth tracking of joint movements, maintaining close alignment between the reference and measured trajectories. Although the trajectories generated from the OCP were stabilized solely in joint space, we plan to implement whole-body stabilization in cartesian space in our future work. Whole-body stabilization in cartesian space would allow for more coordinated control of the exoskeleton, ensuring stability not only in the joints but across the entire body. This would improve the interaction with the environment, especially in dynamic tasks, enhancing the overall robustness of the exoskeleton's performance. For brevity due to space limitations, the graphical results of sitting and standing with a human are not included.

\vspace{-0.4cm}

\section{Conclusion and Outlook}
\label{sec_conclusion}
This study demonstrates the successful execution of sitting, standing and static walking movements on a series-parallel hybrid Recupera-Reha exoskeleton. Exploiting the DDP-based OC algorithm to generate feasible movements highlights the potential for enhancing the exoskeletons' capabilities using OC methods. Additionally, the HyRoDyn solver efficiently handles loop closure constraints in the exoskeleton model and enables a smooth mapping between the independent joint and actuator spaces. 
%Both simulation and experimental results highlight the potential of using the OC approach to generate feasible motions for the exoskeleton, mark the exoskeleton's first walking instance using the DDP-based OC algorithm.
%
% ADD A DISCUSSION HERE, 4 points:
%
% 1- use closed loops in OC formalization
% 2- Improve the design of the exo, adding specific attention to the lower part as well.
% 3- gear ratio transparency through the use of direct drives in a future robot will enable better torque control
% 4- Model human-robot interaction by including it, for example, as part of the closed-loop constraints in the formalization of the OC problem.
In this work, the formalization of OCPs was achieved using the tree abstraction model, due to the high complexity of our system involving a total of 102 constraints. 
Including closed-loop constraints in the optimization process is an effective way to explore the full capabilities of a series-parallel hybrid robot as recently reported in~\cite{boukheddimi2023inv}.
However, the inclusion of such a large number of holonomic constraints makes the resolution of OCP very challenging. In our future work, we will develop new techniques to include the large number of closed-loop constraints in the optimization problem in a computationally efficient manner such that the extreme capabilities of the system can be exploited.
%The system was originally designed for upper-body rehabilitation and has not been optimized for lower-limb rehabilitation, making this work more challenging.
In the future, we would like to optimize the exoskeleton's design for whole-body control in rehabilitation, broadening its potential applications.
Additionally, motors used in the system have a high gear ratio and hence the drives are not mechanically transparent.
We would like to investigate the design of a lightweight exoskeleton robot with high-power quasi-direct drives for precise torque control in the future.
The experiment with the human wearing the exoskeleton was carried out without modeling of the human-robot connections, as the formulation of these connections is not yet mature enough in the literature.
Extending this work, we plan to model these human-robot connections with closed-loop constraints and include them in the trajectory optimization process involving the presence of the human in the decision process.
This methodology could open up a new avenue for the assist-as-needed control of exoskeletons, giving them the ability to perform highly dynamic movements with a human in the loop, in a safe and efficient manner.

\bibliographystyle{IEEEtran}
\bibliography{references}

\begin{thebibliography}{10}
\providecommand{\url}[1]{#1}
\csname url@rmstyle\endcsname
\providecommand{\newblock}{\relax}
\providecommand{\bibinfo}[2]{#2}
\providecommand\BIBentrySTDinterwordspacing{\spaceskip=0pt\relax}
\providecommand\BIBentryALTinterwordstretchfactor{4}
\providecommand\BIBentryALTinterwordspacing{\spaceskip=\fontdimen2\font plus
\BIBentryALTinterwordstretchfactor\fontdimen3\font minus
  \fontdimen4\font\relax}
\providecommand\BIBforeignlanguage[2]{{%
\expandafter\ifx\csname l@#1\endcsname\relax
\typeout{** WARNING: IEEEtran.bst: No hyphenation pattern has been}%
\typeout{** loaded for the language `#1'. Using the pattern for}%
\typeout{** the default language instead.}%
\else
\language=\csname l@#1\endcsname
\fi
#2}}

\bibitem{gorgey2018robotic}
A.~Gorgey, ``Robotic exoskeletons: The current pros and cons,'' \emph{World
  Journal of Orthopedics}, vol.~9, pp. 112--119, 09 2018.

\bibitem{siobhan2021exo}
S.~O'Connor, ``Exoskeletons in nursing and healthcare: A bionic future,''
  \emph{Clinical Nursing Research}, vol.~30, 08 2021.

\bibitem{tijjani2022survey}
I.~Tijjani, S.~Kumar, and M.~Boukheddimi, ``A survey on design and control of
  lower extremity exoskeletons for bipedal walking,'' \emph{Applied Sciences},
  vol.~12, no.~5, p. 2395, 2022.

\bibitem{kumar2020survey}
S.~Kumar, H.~W{\"o}hrle, J.~de~Gea~Fern{\'a}ndez, A.~M{\"u}ller, and
  F.~Kirchner, ``A survey on modularity and distributivity in series-parallel
  hybrid robots,'' vol.~68, 2020.

\bibitem{Felis2016rbdl}
M.~L. Felis, ``Rbdl: an efficient rigid-body dynamics library using recursive
  algorithms,'' \emph{Autonomous Robots}, pp. 1--17, 2016.

\bibitem{delp2007opensim}
S.~L. Delp, F.~C. Anderson, A.~S. Arnold, P.~Loan, A.~Habib, C.~T. John,
  E.~Guendelman, and D.~G. Thelen, ``Opensim: Open-source software to create
  and analyze dynamic simulations of movement,'' \emph{IEEE Transactions on
  Biomedical Engineering}, vol.~54, no.~11, pp. 1940--1950, 2007.

\bibitem{carpentier2019pinocchio}
J.~Carpentier, G.~Saurel, G.~Buondonno, J.~Mirabel, F.~Lamiraux, O.~Stasse, and
  N.~Mansard, ``The pinocchio c++ library -- a fast and flexible implementation
  of rigid body dynamics algorithms and their analytical derivatives,'' in
  \emph{SII}.\hskip 1em plus 0.5em minus 0.4em\relax IEEE, 2019.

\bibitem{meng2023concept}
Q.~Meng, B.~Kong, Q.~Zeng, C.~Fei, and H.~Yu, ``Concept design of
  hybrid-actuated lower limb exoskeleton to reduce the metabolic cost of
  walking with heavy loads,'' vol.~18, no.~5, pp. 1--19, 05 2023.

\bibitem{maryam2019human}
M.~Khamar, M.~Edrisi, and M.~Zahiri, ``Human-exoskeleton control simulation,
  kinetic and kinematic modeling and parameters extraction,'' \emph{MethodsX},
  vol.~6, pp. 1838--1846, 2019.

\bibitem{harib2018feedback}
O.~Harib, A.~Hereid, A.~Agrawal, T.~Gurriet, S.~Finet, G.~Boeris, A.~Duburcq,
  M.~E. Mungai, M.~Masselin, A.~D. Ames, \emph{et~al.}, ``Feedback control of
  an exoskeleton for paraplegics: Toward robustly stable, hands-free dynamic
  walking,'' \emph{IEEE Control Systems Magazine}, vol.~38, no.~6, pp. 61--87,
  2018.

\bibitem{2017_kumar_mimic}
\BIBentryALTinterwordspacing
S.~Kumar, M.~Simnofske, B.~Bongardt, A.~M\"{u}ller, and F.~Kirchner,
  ``Integrating mimic joints into dynamics algorithms: Exemplified by the
  hybrid recupera exoskeleton,'' in \emph{Proceedings of the 2017 3rd
  International Conference on Advances in Robotics}, ser. AIR '17.\hskip 1em
  plus 0.5em minus 0.4em\relax New York, NY, USA: Association for Computing
  Machinery, 2017. [Online]. Available:
  \url{https://doi.org/10.1145/3132446.3134891}
\BIBentrySTDinterwordspacing

\bibitem{Kum18}
S.~Kumar, K.~A.~v. Szadkowski, A.~Mueller, and F.~Kirchner, ``An analytical and
  modular software workbench for solving kinematics and dynamics of
  series-parallel hybrid robots,'' \emph{Journal of Mechanisms and Robotics},
  vol.~12, no.~2, 02 2020, 021114.

\bibitem{kumar2022modular}
R.~Kumar, S.~Kumar, A.~M{\"u}ller, and F.~Kirchner, ``Modular and hybrid
  numerical-analytical approach - a case study on improving computational
  efficiency for series-parallel hybrid robots,'' in \emph{2022 IEEE/RSJ
  International Conference on Intelligent Robots and Systems (IROS)}, 2022, pp.
  3476--3483.

\bibitem{andreas2022dynamics}
A.~M{\"u}ller, ``Dynamics of parallel manipulators with hybrid complex limbs
  — modular modeling and parallel computing,'' \emph{Mechanism and Machine
  Theory}, vol. 167, p. 104549, 2022.

\bibitem{tijjani2022finding}
I.~Tijjani, ``Finding optimal placement of the almost spherical parallel
  mechanism in the recupera-reha lower extremity exoskeleton for enhanced
  workspace,'' in \emph{Advances in Service and Industrial Robotics},
  A.~M{\"u}ller and M.~Brandst{\"o}tter, Eds.\hskip 1em plus 0.5em minus
  0.4em\relax Cham: Springer International Publishing, 2022, pp. 536--544.

\bibitem{Chaichaowarat2023transformable}
R.~Chaichaowarat, S.~Prakthong, and S.~Thitipankul, ``Transformable wheelchair
  exoskeleton hybrid robot for assisting human locomotion,'' \emph{Robotics},
  vol.~12, no.~1, 2023.

\bibitem{tian2024self}
D.~Tian, W.~Li, J.~Li, F.~Li, Z.~Chen, Y.~He, J.~Sun, and X.~Wu,
  ``Self-balancing exoskeleton robots designed to facilitate multiple
  rehabilitation training movements,'' \emph{IEEE Transactions on Neural
  Systems and Rehabilitation Engineering}, vol.~PP, pp. 1--1, 01 2024.

\bibitem{yu2023design}
L.~Yu, H.~Leto, and S.~Bai, ``Design and gait control of an active lower limb
  exoskeleton for walking assistance,'' \emph{Machines}, vol.~11, no.~9, 2023.

\bibitem{Shahrokhshahi2022sample}
A.~Shahrokhshahi, M.~Khadiv, A.~Taherifar, S.~Mansouri, E.~J. Park, and
  S.~Arzanpour, ``Sample-efficient policy adaptation for exoskeletons under
  variations in the users and the environment,'' \emph{IEEE Robotics and
  Automation Letters}, vol.~7, no.~4, pp. 9020--9027, 2022.

\bibitem{featherstone2014rigid}
R.~Featherstone, \emph{Rigid body dynamics algorithms}.\hskip 1em plus 0.5em
  minus 0.4em\relax Springer, 2014.

\bibitem{stewart1965a}
D.~Stewart, ``A platform with six dof,'' \emph{Proceedings of the Institution
  of Mechanical Engineers}, vol. 180, no.~1, pp. 371--386, 1965.

\bibitem{Kumar2018design}
S.~Kumar, B.~Bongardt, M.~Simnofske, and F.~Kirchner, ``Design and kinematic
  analysis of the novel almost spherical parallel mechanism active ankle,''
  \emph{Journal of Intelligent \& Robotic Systems}, vol.~94, pp. 303--325, 05
  2019.

\bibitem{Die17}
M.~Diehl and S.~Gros, ``Numerical optimal control (preliminary and incomplete
  draft),'' 2017.

\bibitem{Bud18}
R.~Budhiraja, J.~Carpentier, C.~Mastalli, and N.~Mansard, ``{Differential
  Dynamic Programming for Multi-Phase Rigid Contact Dynamics},'' in
  \emph{{IEEE-RAS Humanoids 2018}}, Beijing, China, Nov 2018.

\bibitem{Man19}
\BIBentryALTinterwordspacing
N.~Mansard, ``Feasibility-prone differential dynamic programming is ddp a
  multiple shooting algorithm?),'' Online, 2019. [Online]. Available:
  \url{https://gepgitlab.laas.fr/loco-3d/crocoddyl}
\BIBentrySTDinterwordspacing

\bibitem{Mas20}
C.~Mastalli, R.~Budhiraja, W.~Merkt, G.~Saurel, B.~Hammoud, M.~Naveau,
  J.~Carpentier, L.~Righetti, S.~Vijayakumar, and N.~Mansard, ``{Crocoddyl: An
  Efficient and Versatile Framework for Multi-Contact Optimal Control},'' May
  2020.

\bibitem{li2016com}
C.-T. Li, Y.-T. Peng, Y.-T. Tseng, Y.-N. Chen, and K.-H. Tsai, ``Comparing the
  effects of different dynamic sitting strategies in wheelchair seating on
  lumbar-pelvic angle,'' \emph{BMC Musculoskeletal Disorders}, vol.~17, 12
  2016.

\bibitem{boukheddimi2023inv}
M.~Boukheddimi, R.~Kumar, S.~Kumar, J.~Carpentier, and F.~Kirchner,
  ``Investigations into exploiting the full capabilities of a series-parallel
  hybrid humanoid using whole body trajectory optimization,'' in \emph{2023
  IEEE/RSJ International Conference on Intelligent Robots and Systems (IROS)},
  2023, pp. 10\,433--10\,439.

\end{thebibliography}
\end{document}